\documentclass[letterpaper]{article} 
\usepackage{aaai25}  
\usepackage{times}  
\usepackage{helvet}  
\usepackage{courier}  
\usepackage[hyphens]{url}  
\usepackage{graphicx} 
\usepackage{natbib}  
\usepackage{caption} 
\usepackage{algorithm}
\usepackage{amsmath,amsfonts}
\usepackage{colortbl}
\usepackage[table,xcdraw]{xcolor}
\usepackage{algorithm}
\usepackage{algpseudocode}
\usepackage{lineno}
\usepackage{bm}
\usepackage{float}
\usepackage{multirow}
\usepackage{makecell}
\usepackage{tabularx}
\usepackage{amssymb}
\usepackage{stmaryrd}
\usepackage{lipsum}
\usepackage{subfigure}
\usepackage{multicol}
\usepackage{bbold}
\usepackage{amsthm}
\usepackage{newfloat}
\usepackage{listings}
\usepackage{tikz}
\usepackage{graphicx}
\usepackage{mathtools}
\usepackage{textcomp}
\usepackage[skins]{tcolorbox}
\tcbuselibrary{breakable}
\usepackage{lipsum} 
\usepackage{xcolor}
\newtheorem{theorem}{Theorem}

\newtheorem{corollary}[theorem]{Corollary}

\usepackage{newfloat}
\usepackage{listings}
\DeclareCaptionStyle{ruled}{labelfont=normalfont,labelsep=colon,strut=off} 
\floatstyle{ruled}
\newfloat{listing}{tb}{lst}{}
\floatname{listing}{Listing}
\title{Bootstrapping Heterogeneous Graph Representation Learning via Large Language Models: A Generalized Approach}
\author{
    Hang Gao\equalcontrib\textsuperscript{\rm 1}
    Chenhao Zhang\equalcontrib\textsuperscript{\rm 1, \rm 2},
    Fengge Wu\textsuperscript{\rm 1, \rm 2}\thanks{Corresponding author.},
    Changwen Zheng\textsuperscript{\rm 1, \rm 2},
    Junsuo Zhao\textsuperscript{\rm 1, \rm 2},\\
    Huaping Liu\textsuperscript{\rm 3}
}
\affiliations{
    \textsuperscript{\rm 1}National Key Laboratory of Space Integrated Information System, Institute of Software, Chinese Academy of Sciences.\\
    \textsuperscript{\rm 2}University of Chinese Academy of Sciences.\\
    \textsuperscript{\rm 3}Tsinghua University. \\
    \{gaohang, zhangchenhao2024, fengge, changwen, junsuo\}@iscas.ac.cn, \\
    hpliu@tsinghua.edu.cn
}

\usepackage{bibentry}

\begin{document}

\maketitle

\begin{abstract}
Graph representation learning methods are highly effective in handling complex non-Euclidean data by capturing intricate relationships and features within graph structures. However, traditional methods face challenges when dealing with heterogeneous graphs that contain various types of nodes and edges due to the diverse sources and complex nature of the data. Existing Heterogeneous Graph Neural Networks (HGNNs) have shown promising results but require prior knowledge of node and edge types and unified node feature formats, which limits their applicability. Recent advancements in graph representation learning using Large Language Models (LLMs) offer new solutions by integrating LLMs' data processing capabilities, enabling the alignment of various graph representations. Nevertheless, these methods often overlook heterogeneous graph data and require extensive preprocessing. To address these limitations, we propose a novel method which leverages the strengths of both LLM and GNN, allowing for the processing of graph data with any format and type of nodes and edges without the need for type information or special preprocessing. Our method employs LLM to automatically summarize and classify different data formats and types, aligns node features, and uses a specialized GNN for targeted learning, thus obtaining effective graph representations for downstream tasks. Theoretical analysis and experimental validation have demonstrated the effectiveness of our method. 
\end{abstract}

\begin{links}
\link{Code, Datasets and Appendix}{https://github.com/zch65458525/GHGRL/tree/main}
\end{links}

%

\section{Introduction}

Graph representation learning methods are highly effective for processing complex non-Euclidean data, as they can model intricate relationships within graph structures. However, real-world scenarios often involve heterogeneous graph data, which consists of various types of nodes and edges due to the diverse sources and complexity of the data \cite{DBLP:journals/tbd/WangBSFYY23}. Examples include social network analysis \cite{DBLP:conf/kdd/QiuTMDW018, li-goldwasser-2019-encoding}, recommendation systems \cite{DBLP:conf/www/Fan0LHZTY19,DBLP:journals/corr/YangYHGD14a}, and traffic prediction \cite{DBLP:conf/aaai/GuoLFSW19}. General graph representation learning methods often struggle to handle this heterogeneity. Therefore, developing methods that can effectively process and learn from graphs with diverse node and edge types is essential to broaden the applicability of graph representation learning and enhance its capability to manage complex data.

To overcome these difficulties, Heterogeneous Graph Neural Networks (HGNNs) have been developed and shown promising results \cite{DBLP:conf/aaai/HongGLYLY20,DBLP:conf/kdd/DongCS17,DBLP:journals/tkde/YangXZSH22}. HGNNs are designed to process graphs with varying node and edge types using specialized techniques, including both metapath-based \cite{DBLP:conf/www/WangJSWYCY19,DBLP:conf/www/0004ZMK20} and metapath-free approaches \cite{DBLP:conf/kdd/FanZHSHML19}. These works leverage meta-path-based aggregation, attention mechanisms, and embedding techniques to effectively manage the diversity of nodes and edges, enabling the processing of heterogeneous graph data. However, HGNNs have limitations that restrict their applicability in scenarios where prior knowledge of node and edge types or consistent node feature formats is unavailable. For example, in open-source intelligence analysis, IoT log analysis, or monitoring malicious internet activities, the unpredictable, diverse, and dynamic nature of the data poses significant challenges in identifying and labeling node types.

\begin{figure}[ht]
    \centering
    \subfigure[The left side of the figure shows the form of input graph data for HGNN, where nodes of different colors represent different types of heterogeneous nodes. The labels for node type and edge type in the graph indicate the required type information for the input data. The right side outlines its characteristics.]{%
        \includegraphics[width=0.48\textwidth]{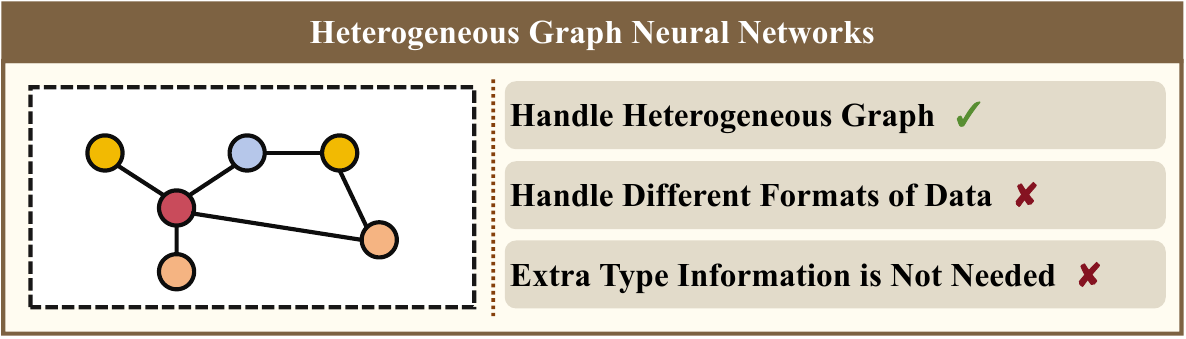}
        \label{fig:intsubfig1}
    }
    \subfigure[Similar to the above figure, the left side of the figure shows the form of input graph data for LLM-based graph representation learning, where nodes of different shapes represent different forms of node attributes.]{%
        \includegraphics[width=0.48\textwidth]{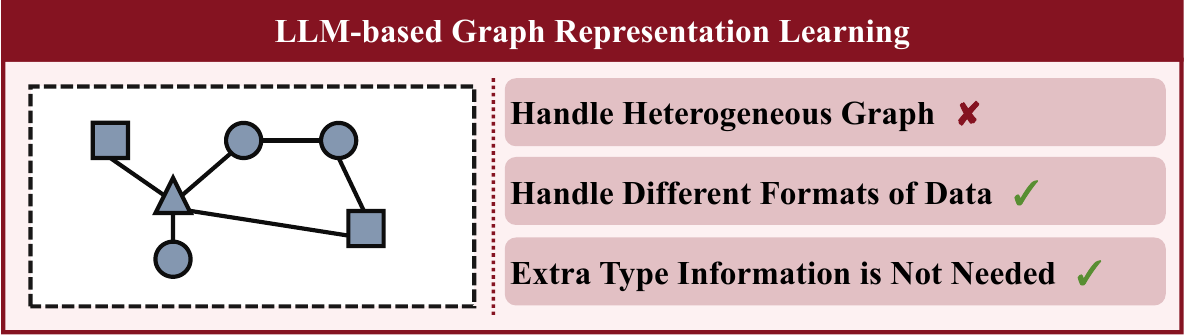}
        \label{fig:intsubfig2}
    }
    \subfigure[Demonstration and summary of our method, following the same demonstration format as the two figures above.]{%
        \includegraphics[width=0.48\textwidth]{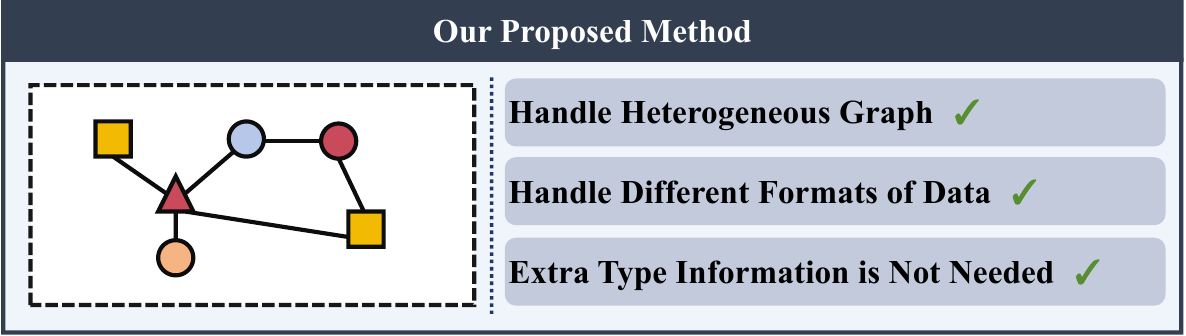}
        \label{fig:intsubfig3}
    }
    \vskip -0.1in
    \caption{Demonstration of different methods.}
    \vskip -0.15in
    \label{fig:intoverall}
\end{figure}

Recently, the emergence of graph representation learning methods based on LLMs \cite{DBLP:conf/naacl/DevlinCLT19, DBLP:journals/corr/abs-2005-14165} has provided new solutions to the aforementioned problems. These methods integrate the background knowledge and data processing capabilities of LLMs into graph representation learning, allowing for the alignment of various types of graph representations based on LLMs \cite{DBLP:journals/sigkdd/ChenMLJWWWYFLT23, DBLP:conf/nips/HuangRCK0LL23}. These approaches can handle diverse graph data, achieving significant results in the field of graph representation learning and providing directions for building foundational models in this area. However, these methods primarily focus on handling different types of homogeneous graph representation learning tasks and overlook the importance of processing heterogeneous graph data. Nevertheless, an effective method capable of processing such data without the need for additional data cleaning and annotation is highly necessary. At the same time, these methods often require a certain degree of preprocessing of the graph data, which limits their practical application. Figure \ref{fig:intsubfig1} and \ref{fig:intsubfig2} provide a practical illustration of these methods.

To address these challenges, we propose a novel \textit{\textbf{G}eneralized \textbf{H}eterogeneous \textbf{G}raph \textbf{R}epresentation \textbf{L}earning} (GHGRL) method. GHGRL integrates the strengths of both LLMs and GNNs to process graph data in a more generalized manner. As demonstrated in Figure \ref{fig:intsubfig3}, GHGRL can handle graph data with nodes and edges of any format and type, without requiring explicit type information or special pre-processing of the data. Specifically, GHGRL utilizes LLMs to process the training data by automatically summarizing and classifying the various data formats and types present in the graph. Subsequently, LLMs are used to align node features across different formats, generating representation vectors for node attributes. Next, we employ our specially designed GNN to perform targeted learning on the graph data based on the types and estimations derived from the LLM, thereby obtaining graph representations suitable for downstream tasks.

\textbf{Contributions:}
\begin{itemize}
\item We propose a novel method that combines LLM and GNN to process heterogeneous graph data without requiring node and edge type information. Additionally, this method can handle scenarios where node attributes are not uniform.

\item We present the specific implementation of the aforementioned method and conduct theoretical analysis and validation of its performance.

\item We developed more challenging datasets to rigorously test the proposed method. Additionally, we validated our approach using widely adopted heterogeneous graph datasets to ensure robustness and reliability.

\end{itemize}

\section{Related Works}
\paragraph{Heterogeneous Graph Representation Learning.} 

Heterogeneous graph representation learning methods are categorized into metapath-based and metapath-free approaches. Metapath-based methods use heterogeneous graph neural networks to aggregate and integrate semantic features \cite{DBLP:conf/nips/YunJKKK19, zhang2019heterogeneous, DBLP:conf/www/WangJSWYCY19, DBLP:conf/www/0004ZMK20, DBLP:journals/air/BingYZMMQ23, yang2023simple}. HetGNN \cite{zhang2019heterogeneous} uses random walks and node type aggregation. HAN \cite{DBLP:conf/www/WangJSWYCY19} and MAGNN \cite{DBLP:conf/www/0004ZMK20} use metapaths for semantic differentiation and propagation. SeHGNN \cite{yang2023simple} extends receptive fields with long metapaths and transformer-based modules. Metapath-free methods embed semantic information using attention mechanisms \cite{zhu2019relation, DBLP:conf/kdd/FanZHSHML19, DBLP:conf/aaai/HongGLYLY20, DBLP:conf/kdd/LvDLCFHZJDT21, DBLP:conf/icml/ZhouSYZ023, DBLP:conf/www/HeWFHLSY24}. HGB \cite{DBLP:conf/kdd/LvDLCFHZJDT21} uses a multi-layer GAT network for node distinction. PSHGCN \cite{DBLP:conf/www/HeWFHLSY24} uses positive spectral heterogeneous graph convolution to learn valid heterogeneous graph filters. These methods all require prior knowledge of node and edge types and are typically used on datasets where these types are known. However, this limitation restricts the application of these methods in the broader field of data mining.

\paragraph{LLMs for Graphs.}
With the emergence of various LLM methods, the use of LLMs for graph representation learning is becoming a research hotspot. Relevant studies can be classified into two types. One type enriches node representation based on prompt learning and processes graph data tasks using GNN \cite{DBLP:journals/corr/abs-2310-04560, DBLP:journals/sigkdd/ChenMLJWWWYFLT23, DBLP:conf/nips/HuangRCK0LL23, DBLP:conf/www/0001RTYC24, liuone, DBLP:conf/sigir/Tang00SSCY024}. The other type converts graph data into text for LLM processing \cite{DBLP:journals/corr/abs-2312-10372,DBLP:conf/nips/WangFHTHT23,DBLP:journals/corr/abs-2305-15066,DBLP:conf/kdd/SunCLLG23,DBLP:journals/corr/abs-2403-04483, DBLP:journals/corr/abs-2403-04780}. These methods handle homogeneous data and require pre-standardized representation of node information, limiting their application in data mining. To address this, our proposed method is designed to integrate LLMs to handle heterogeneous graph attributes of any format and type without prior knowledge, expanding the application scope in data mining. Please refer to \textbf{Appendix} A for an extended related work.

\section{Methodology}

\begin{figure*}[h]
  \centering
  \includegraphics[width=1\linewidth]{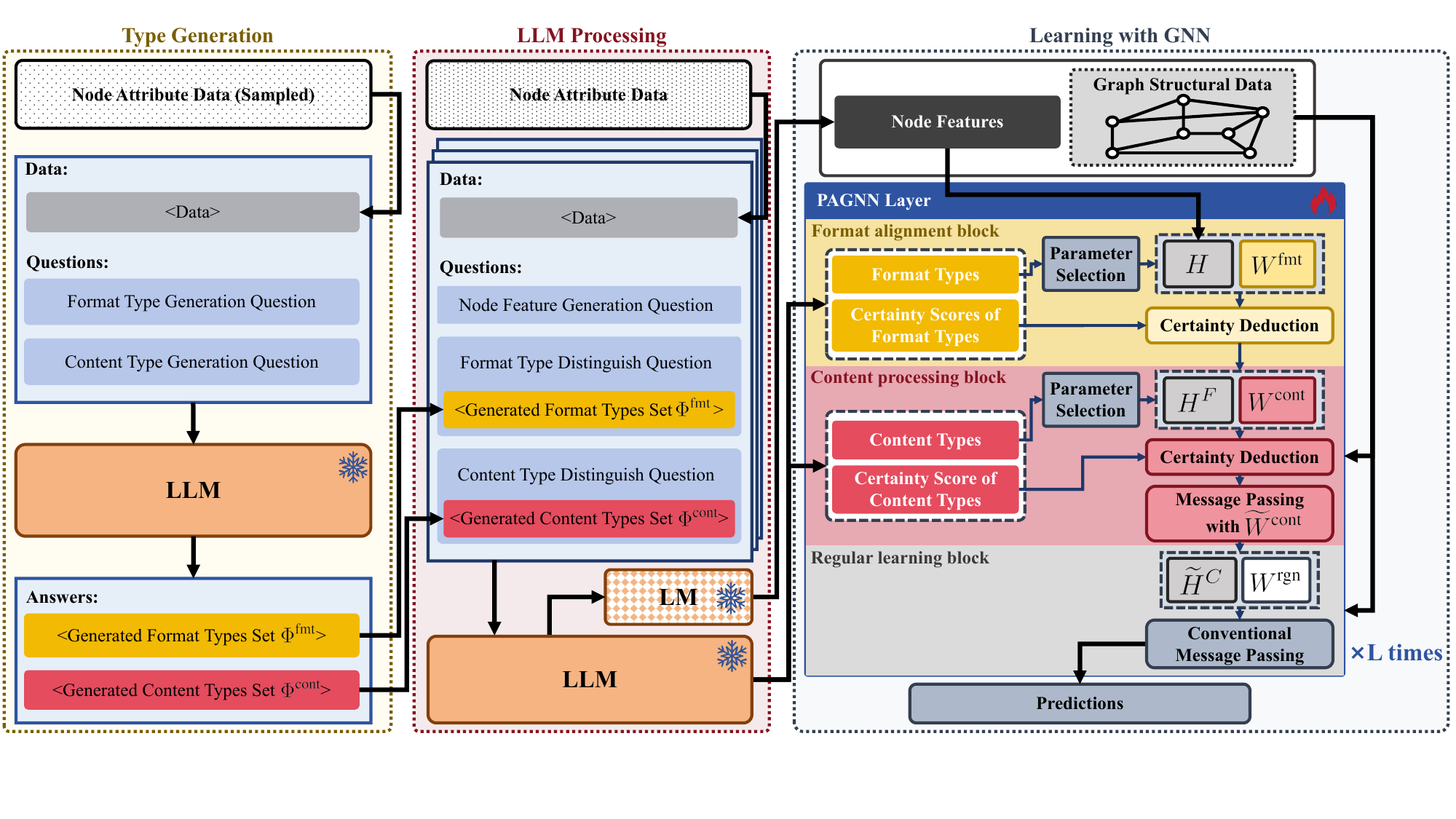}
  \caption{The framework of the proposed method. The snowflake symbol represents the fixed model parameters, while the flame represents the model parameters involved in training.}
    \vskip -0.05in
  \label{fig:framework}
  \vskip -0.1in
\end{figure*}

Our proposed GHGRL framework aims to enhance learning on heterogeneous graphs using LLMs, thereby making the processing capabilities more generalized. Specifically, for a heterogeneous graph $G = \{\mathcal{V}, \mathcal{E}\}$, $G$ contains various types of nodes with different representation formats. Additionally, the edges between adjacent nodes may also possess different types. Moreover, the types of nodes and edges are unknown to us. Our goal is to construct a model that can effectively handle $G$ and accomplish graph representation learning tasks. To achieve this, GHGRL is divided into the following three modules: \textbf{1) Type Generation}, which identifies all possible node types based on node attributes; \textbf{2) LLM Processing}, which estimates the specific type of each node and generates node representation vectors using LLMs; and \textbf{3) Learning with GNN}, which leverages the acquired node types and representations for graph learning with a specially designed GNN. During the GNN learning process, different types of edges are distinguished, and message passing is executed accordingly. Figure \ref{fig:framework} demonstrates the framework of GHGRL.

\subsection{Type Generation}

First, since we do not have access to the number of node types in the dataset or detailed information about them, we opt to generate these types directly. We create two categories of node type set: the format-based set $\Phi^{\text{fmt}}$ and the content-based set $\Phi^{\text{cont}}$. Specifically, we randomly select a subset of node attribute samples, denoted as $\bm{\tilde{X}} = \{ \bm{x}_{i}\}_{i=1}^{|\widetilde{X}|}$, from the training set and input them into the LLM, allowing it to analyze and identify the node types present in the dataset. We use Llama 3 \cite{dubey2024llama3herdmodels} as our backbone LLM. The number of selected samples, $|\widetilde{X}|$, is determined by the maximum input sequence length of the LLM.

As a result, we obtain the generated format types set $\Phi^{\text{fmt}} = \{s^{\text{fmt}}_{j}\}_{j=1}^{m^{\text{fmt}}}$, where each $s^{\text{fmt}}_{j}$ represents a generated string-formatted node format type name, represented in text format, such as ``noun'' or ``detail description''. Similarly, the generated content types set is $\Phi^{\text{cont}} = \{s^{\text{cont}}_{j}\}_{j=1}^{m^{\text{cont}}}$, where each $s^{\text{cont}}_{j}$ denotes a generated string-formatted node content type name, also represented in text format, such as ``paper concerning deep learning'' or ``paper concerning biology''. The parameters $m^{\text{fmt}}$ and $m^{\text{cont}}$ are hyperparameters controlling the size of $\Phi^{\text{fmt}}$ and $\Phi^{\text{cont}}$, respectively. Formally, we have:
\begin{equation}
    \left\{ \Phi^{\text{fmt}}, \Phi^{\text{cont}} \right\} = LLM\left(\bm{\tilde{X}},m^{\text{fmt}},m^{\text{cont}},P^{G}\right),
\end{equation}
where $P^{G}$ denotes the type generation prompt. Please refer to \textbf{Appendix} C for details.

\subsection{LLM Processing}
Subsequently, we process the data with the LLM to acquire node features, estimating the format type and content type of each node's attribute features. Based on $\Phi^{\text{fmt}}$ and $\Phi^{\text{cont}}$, we conduct analysis upon each node $v$'s feature $\bm{x}_{v}$. We obtain five different outputs: description text $\bm{h}^{\text{desc}}_{v}$ of node $v$, format type estimation result $\phi^{\text{fmt}}(v)$, format type estimation confidence score $c^{\text{fmt}}(v)$, content type estimation result $\phi^{\text{cont}}(v)$, content type estimation confidence score $c^{\text{cont}}(v)$, description text $\bm{h}^{\text{reas}}_{v}$ of the estimation reasons. $\phi^{\text{fmt}}(v)$ denotes the index of the estimated format type of node $v$ within $\Phi^{\text{fmt}}$, while $\phi^{\text{cont}}(v)$ denotes the index of the estimated content type of node $v$ within $\Phi^{\text{cont}}$. Formally, we have:
\begin{gather}
    \left\{\bm{h}^{\text{desc}}_{v}, \phi^{\text{fmt}}(v), c^{\text{fmt}}(v), \phi^{\text{cont}}(v), c^{\text{cont}}(v), \bm{h}^{\text{reas}}_{v}\right\} = 
    \nonumber\\
    LLM\left(\bm{x}_{v}, \Phi^{\text{fmt}}, \Phi^{\text{cont}}, P^{P}\right).
\end{gather}

We ensure that the LLM outputs as much information about the node attributes as possible by modifying the prompts. Please refer to \textbf{Appendix} C for details. In addition to node descriptions, we also require the model to provide a reasoning description for its estimation of the node types. This approach allows the model to refine and optimize the modeling of heterogeneous graphs.

Subsequently, we adopt a language model sentence transformer \cite{DBLP:conf/emnlp/ReimersG19} to generate fixed-length node representation based on both $\bm{h}^{\text{desc}}_{v}$ and $\bm{h}^{\text{reas}}_{v}$, formally, we have:
\begin{gather}
    \bm{h}_{v} = s(\bm{h}^{\text{desc}}_{v}, \bm{h}^{\text{reas}}_{v}),
\end{gather}
where $s(\cdot)$ denotes the sentence transformer model. $\bm{h}_{v}$ will be utilized as the node feature in the subsequent modules.

\subsection{Learning with GNN}

To integrate LLM estimates into graph representation learning, we specifically designed a novel \textit{\textbf{P}arameter \textbf{A}daptive \textbf{G}NN} (PAGNN) to maximize the utilization of LLM outputs for graph data processing. Part of the PAGNN structure is determined by the LLM outputs. Specifically, each layer of PAGNN includes three components: a format alignment block based on format type, a heterogeneous processing block based on content type, and a regular learning block. These components will be introduced in detail below.

\paragraph{Format alignment block.} The purpose of this block is to align node features represented in different forms. This block utilizes matrix $W^{\text{fmt}} \in \mathbb{R}^{ |\Phi^{\text{fmt}}| \times d^{\text{in}} \times d^{\text{fmt}}  }$ and $B^{\text{fmt}} \in \mathbb{R}^{ |\Phi^{\text{fmt}}| \times d^{\text{fmt}} }$ as the network parameters, where $d^{\text{in}}$ and $d^{\text{fmt}}$ denote the input and output feature width of this block, $|\Phi^{\text{fmt}}|$ is the number of format types. Subsequently, with the input node representation matrix $H$, where $H \in \mathbb{R}^{ |\mathcal{V}| \times x^{\text{fmt}} } $, $|\mathcal{V}|$ is the number of nodes, for all $v \in \{1,2,...,|\mathcal{V}|\}$ the block performs the following calculation:
\begin{gather}
    H^{\text{fmt}  \ [v]} = \delta\left(  H^{[v]}   W^{\text{fmt} \ [\phi^{\text{fmt}}(v)]} + B^{\text{fmt} \ [\phi^{\text{fmt}}(v)]} \right), 
    \label{eq:fab}
\end{gather}
where \( W^{\text{fmt} \ [\phi^{\text{fmt}}(v)]} \) and \( B^{\text{fmt} \ [\phi^{\text{fmt}}(v)]} \) denote the \(\phi^{\text{fmt}}(v)\)-th elements of \( W^{\text{fmt}} \) and \( B^{\text{fmt}} \) along the first dimension, respectively. $H^{\text{fmt}  \ [v]}$ denotes the $v$-th vector within $H^{\text{fmt}}$. This design ensures that all nodes with the same type utilize the same set of parameters, while nodes with different types utilize different sets of parameters. Due to the potential inaccuracy and possible misjudgment of node type estimation by LLM, we further introduce the generated confidence score $c^{\text{fmt}}(v)$ and adjust the blocks based on this score. We optimize Equation \ref{eq:fab} to generate a formal representation of the block with the added confidence score as follows:
\begin{gather}
        H^{\text{fmt} \  [v]} = \delta\Big(  c^{\text{fmt}}\left(v\right)  \big( H^{[v]}   W^{\text{fmt} \ [\phi^{\text{fmt}}\left(v\right)]} + B^{\text{fmt} \ [\phi^{\text{fmt}}(v)]} \big) \  +
        \nonumber \\
        \left(1-c^{\text{fmt}}\left(v\right)\right) H^{[v]}\Big), 
\end{gather}
where $c^{\text{fmt}}(v)$ is the aforementioned format type confidence score of $v$. $\delta(\cdot)$ denotes the activate function. This design ensures that the effect of $W^{\text{fmt} \ [\phi^{\text{fmt}}\left(v\right)]}$ decreases as the confidence score decreases.

\paragraph{Content processing block.} This block trails the format alignment block. It processes node features of different generated node content types and then conducts message passing between them. The pattern of information transmission between different nodes may vary. Conventional heterogeneous graph representation learning methods often use meta paths or predefined edge types to address this issue \cite{DBLP:conf/www/WangJSWYCY19,DBLP:conf/www/0004ZMK20,yang2023simple}. However, since we cannot obtain this information, we can only differentiate information transmission based on the node content type generated by the LLM. Specifically, the content processing block first conducts the following calculation:
\begin{gather}
            H^{\text{cont} \ [v]} = \delta\Big( c^{\text{cont}}(v) \big( H^{\text{fmt} \ [v]}   W^{\text{cont} \ [\phi^{\text{cont}}(v)]} 
            \nonumber\\
            + \ B^{\text{cont} \ [\phi^{\text{cont}}(v)]} \big) \Big), 
            \label{eq:prec}
\end{gather}
where $c^{\text{cont} }(v)$ is the aforementioned content type confidence score of $v$, $W^{\text{cont}} \in \mathbb{R}^{ |\Phi^{\text{cont}}| \times d^{\text{fmt}} \times d^{\text{cont}}  }$ and $B^{\text{cont}} \in \mathbb{R}^{ |\Phi^{\text{cont}}| \times d^{\text{cont}} }$ are parameter matrices. $d^{\text{cont}}$ denotes the feature width of each representation vector within $H^{\text{cont} \ [v]}$. The content processing block then conducts message passing with $H^{\text{cont}}$:
\begin{gather}
        \widetilde{H}^{\text{cont} \ [v]} =  \alpha H^{\text{cont} \ [v]} \ \ +   
        \nonumber\\
        AGG\left(H^{\text{cont} \ [u]} \widetilde{W}^{\text{cont} \ [\phi(v)] } ,u \in \mathcal{N}\left(v\right)\right)  ,       
        \label{eq:aftc}
\end{gather}
where $\alpha$ be a hyperparameter to control the proportion of original node features, parameter matrix $\widetilde{W}^{\text{cont}} \in \mathbb{R}^{ |\Phi^{\text{cont}}| \times d^{\text{cont}} \times d^{\text{cont}}  }$. $AGG(\cdot)$ aggregates the features from neighbors. We adopt $d^{\text{cont}}$ again as the output feature width of content processing block. Equation \ref{eq:prec} and \ref{eq:aftc} actually ensure that during the aggregation operation, the node representations are multiplied by the corresponding parameter matrices according to the content type of the source and target nodes of the edges. This, in turn, ensures that the entire data aggregation process maximally distinguishes between different node types and edge types.

\paragraph{Regular learning block.} This block follows the first two blocks and can be formally represented as follows.
\begin{gather}
        H^{\text{rgn} \ [v]} =  \delta\bigg(\widetilde{H}^{\text{cont} \ [v]} W^{\text{rgn}}\ \ +   
        \nonumber\\
        AGG\left(\widetilde{H}^{\text{cont} \ [u]} W^{\text{rgn}} ,u \in \mathcal{N}\left(v\right)\right)\bigg),
        \label{eq:rgn}
\end{gather}
which is similar to a regular GCN layer, adopting the same method for data propagation to learn the common features present in the data. $W^{\text{rgn}} \in \mathbb{R}^{ d^{\text{cont}} \times d^{\text{rgn}}  }$ is the parameter matrix. $d^{\text{rgn}}$ denotes the output feature width of this block.


The aforementioned blocks form a PAGNN layer. Our model is composed of multiple PAGNN layers, and we remove the format alignment block and the content processing block after the $l^{\text{fmt}}$ layer and $l^{\text{cont}}$ layer respectively, as the heterogeneity of node features is sufficiently represented by that point. $l^{\text{fmt}} \leq l^{\text{cont}} \leq L$, where L is the total number of network layers. Please refer to \textbf{Appendix} D for details.

\section{Analysis}
In this section, we further analyze the proposed method by examining how GHGRL effectively learns various types of semantics. This capability helps to mitigate semantic confusion, which can arise from the over-smoothing common in conventional graph representation learning methods. Such over-smoothing can be problematic when dealing with complex heterogeneous graph data \cite{DBLP:conf/icml/ZhouSYZ023}. We adopt a simplified graph convolution model $g(\cdot)$ \cite{DBLP:conf/iclr/KipfW17} for this type of analysis, the layer structure of which can be represented as follows:
\begin{gather}
            \bm{h}_{v}^{(l+1)} = \bm{h}_{v}^{(l)} + AGG\left(\bm{h}_{u}^{(l)}W^{(l)},u \in \mathcal{N}\left(v\right)\right),
            \label{eq:shgnn}
\end{gather}
where $\bm{h}_{v}^{(l+1)}$ and $\bm{h}_{v}^{(l)}$ denote the output representation of node $v$ of layer $l$ and $l+1$ respectively. Several related works \cite{DBLP:conf/icml/ZhouSYZ023, DBLP:conf/aaai/LiHW18} have demonstrated that this model can effectively represent the properties of different types of graph neural networks, making it widely applicable in the analysis of the over-smoothing characteristics of graph neural networks. Furthermore, it bears significant similarity to our model's architecture.

GNN models generally suffer from over-smoothing \cite{DBLP:conf/icml/ChenWHDL20,DBLP:conf/iclr/GeertsR22}. Specifically, for graph convolution model $g(\cdot)$ with $L$ layers and graph $G = \{\mathcal{V}, \mathcal{E}\}$, we can represent the limitation of $g$ when $L \to + \infty$:
\begin{gather}
            \lim_{L \to + \infty} g(\mathcal{V}, \mathcal{E}) = \begin{bmatrix}\bm{h}_{1}^{(L)}& \bm{h}_{2}^{(L)}& \cdots& \bm{h}_{|\mathcal{V}|}^{(L)}\end{bmatrix}^\top,
            \label{eq:os}
\end{gather}
where $\bm{h}_{v}^{(L)}$ denotes the $L$-th layer output representation of node $v$. For any node $i$ and $j$ within $G$, $\bm{h}_{i}^{(L)}$ and $\bm{h}_{j}^{(L)}$ are linearly dependent. Yet, our proposed GHGRL could avoid such over-smoothing. To prove this, we construct a simplified model $\widetilde{g}(\cdot)$ of GHGRL, $l$-th layer of $\widetilde{g}(\cdot)$ possesses the following form:
\begin{gather}
            \bm{h}_{v}^{(l+1)} = \bm{h}_{v}^{(l)} \ \ + 
            \nonumber\\
            AGG\left(\bm{h}_{u}^{(l)}W^{ [\phi(v)] } + B^{ [\phi(v)] } ,u \in \mathcal{N}\left(v\right)\right).
            \label{eq:shgnnex}
\end{gather}
The input of $\widetilde{g}(\cdot)$ are node features extracted from LLM $f(\cdot)$. Subsequently, we propose the following theorem.
\begin{theorem}
    Given a connected graph $G = \{\mathcal{V}, \mathcal{E}\}$ with node features $\{\bm{x}_{i}\}_{i=1}^{|\mathcal{V}|}$ and LLM $f(\cdot)$, $\widetilde{g}(\cdot)$ can avoid the over-smoothing described in Equation \ref{eq:os} for the node features, i.e., we have:
    \begin{equation}
        \lim_{L \to + \infty} \widetilde{g}(\{f(\bm{x}_{j})\}_{j=1}^{|\mathcal{V}|}, \mathcal{E}) = \begin{bmatrix}\Tilde{\bm{h}}_{1}^{(L)}& \Tilde{\bm{h}}_{2}^{(L)}& \cdots& \Tilde{\bm{h}}_{|\mathcal{V}|}^{(L)}\end{bmatrix}^\top,
        \label{eq:dsim}
    \end{equation}
    where for node $i$ and $j$ that satisfying $\phi(i) \neq \phi(j)$, $\Tilde{\bm{h}}_{i}$ and $\Tilde{\bm{h}}_{j}$ are linearly independent.
    \label{thm:m}
\end{theorem}
The proof can be found in \textbf{Appendix} B.1. Theorem \ref{thm:m} demonstrates through the model in Equation \ref{eq:shgnnex} that our method effectively prevents over-smoothing among different types of nodes. This ensures that the model preserves the distinctive features between various node types. Moreover, this differentiation is automatically derived based on the judgments produced by an LLM, ensuring that our model can leverage the knowledge of the LLM for type estimation. Consequently, this facilitates relation learning training based on the structure of the GHGRL.
\begin{corollary}
    Given the conditions in Theorem \ref{thm:m}, if node $i$ and $j$ satisfied $\phi(i) = \phi(j)$, $i$ and $j$ do not share same set of neighbors, then $\Tilde{\bm{h}}^{(L)}_{i}$ and $\Tilde{\bm{h}}^{(L)}_{j}$ are not necessarily linear dependent for $L \to + \infty$.
    \label{thm:cly}
\end{corollary}

The proof can be found in \textbf{Appendix} B.2. Corollary \ref{thm:cly} further demonstrates that our proposed method not only prevents over-smoothing between different types of heterogeneous nodes, but also ensures that over-smoothing does not necessarily occur between nodes of the same type. In such cases, whether over-smoothing occurs depends on the types of adjacent nodes and network parameters. This means the network can adaptively make node features similar or different based on specific circumstances, rather than causing all node features to converge to the same value due to over-smoothing. As shown in \cite{DBLP:conf/aaai/LiHW18}, a 3-layer GCN can already experience over-smoothing on certain datasets. With two aggregations per layer, our 3-layer PAGNN is equivalent to a 6-layer GCN (Please refer to Section Experiments for details), heightening the risk of over-smoothing.  We analyzed inter-type node similarity across layers and compared our method to GHGRL without PAGNN (GHGRL w/o P). The results below confirm that over-smoothing occurs, and our method effectively prevents it. 

\begin{table}[h!]\tiny
    \renewcommand{\arraystretch}{1.1}
    \centering
    \caption{Mean cosine similarity between each type's average feature vector and the overall mean, indicating the degree of over-smoothing on the IMDB dataset.}
    \begin{tabular}{@{}l|cccc@{}}
        \hline
        Method & Layer 1 & Layer 2 & Layer 3 & Layer 4 \\
        \hline
        GHGRL w/o P & -0.151 & 0.327 & 0.702 & 0.882 \\
        GHGRL & -0.133 & 0.174 & 0.311 & 0.395 \\
        \hline
    \end{tabular}
\end{table}

\section{Experiments}

\begin{table*}[ht]\tiny
    \renewcommand{\arraystretch}{1.1}
    \setlength{\tabcolsep}{3.5pt}
    \centering
        \caption{Comparative experiment results for heterogeneous graph datasets. \textbf{Bold} denotes the best performance, \underline{underline} denotes the second best. ``-w'' denotes results of HGNN method without type information. }
        \vskip -0.1in
    \begin{tabular}{l|cccc|cccc|cccc}
        \hline
               Datasets & \multicolumn{2}{c}{IMDB (10\% Training)} & \multicolumn{2}{c|}{IMDB (40\% Training)} &  \multicolumn{2}{c}{DBLP (10\% Training)} &  \multicolumn{2}{c|}{DBLP (40\% Training)} &  \multicolumn{2}{c}{ACM (10\% Training)} &  \multicolumn{2}{c}{ACM  (40\% Training)} \\
        \cline{1-13}
         Metrics & Macro-F1 & Micro-F1 & Macro-F1 & Micro-F1 & Macro-F1 & Micro-F1  & Macro-F1 & Micro-F1  & Macro-F1 & Micro-F1   & Macro-F1 & Micro-F1 \\
        \hline
         GCN   & 57.47$\pm$0.72             &  58.43 $\pm$1.15   & 60.13$\pm$0.76 & 60.38$\pm$1.19           & 89.09$\pm$0.32                    & 89.8$\pm$0.34 & 88.94$\pm$0.38 & 89.61$\pm$0.40 &89.47$\pm$0.23 &90.23$\pm$0.24 &89.19$\pm$0.28 &89.95$\pm$0.27\\
         GAT  & 60.12$\pm$0.79             & 60.79$\pm$1.26                & 62.85$\pm$1.28            & 63.1$\pm$0.83 & 89.66$\pm$0.26           & 90.93$\pm$0.23 &91.40$\pm$0.19 &91.79$\pm$0.21 & 92.23$\pm$0.96 &92.27$\pm$0.95 &92.26$\pm$0.86 &92.38$\pm$0.81\\
         \hline
         HAN  & 61.28$\pm$0.12   & 61.26$\pm$0.15   & 62.78$\pm$0.38    & 62.15$\pm$0.26    & 91.23$\pm$0.51    & 92.10$\pm$0.62    & 91.92$\pm$0.48    & 92.52$\pm$0.59    & 90.58$\pm$0.40    & 90.56$\pm$0.39    & 92.70$\pm$0.45    & 92.75$\pm$0.42\\
         MAGNN  & 57.78$\pm$2.85    & 57.97$\pm$1.82    & 59.92$\pm$1.24    & 60.07$\pm$0.89    & 92.24$\pm$0.49    & 92.70$\pm$0.51 & 93.21$\pm$0.42    & 93.68$\pm$0.43   & 89.46$\pm$0.64    & 89.71$\pm$0.53    & 91.25$\pm$0.24    & 91.33$\pm$0.35\\

        SeHGNN  & 61.23$\pm$0.46   & \underline{62.74$\pm$0.37} & 62.62$\pm$0.35
        & 65.34$\pm$0.30    & \textbf{93.74$\pm$0.28}    & \textbf{94.19$\pm$0.24}    & \textbf{94.48$\pm$0.12}   & \textbf{94.85$\pm$0.15}    & \underline{92.06$\pm$0.32}    & \underline{92.10$\pm$0.32}    & 93.38$\pm$0.30    & 93.44$\pm$0.36\\

        PSHGCN  & \underline{61.35$\pm$0.79}   & 62.25$\pm$0.42 & \underline{67.21$\pm$0.66} & \underline{67.55$\pm$0.56}   & \underline{92.89$\pm$0.09}    & \underline{93.46$\pm$0.07}    & \underline{93.98$\pm$0.12}    & \underline{94.29$\pm$0.10}    & 91.07$\pm$0.26    & 91.00$\pm$0.24    & \underline{93.78$\pm$0.23}    & \underline{93.77$\pm$0.19}\\
        \hline
        \rowcolor{orange!20}
        HAN-w  & 58.31$\pm$0.32   & 58.26$\pm$0.31   & 59.83$\pm$0.33    & 59.02$\pm$0.35    & 87.54$\pm$0.78    & 87.93$\pm$0.58    & 88.16$\pm$0.68    & 88.64$\pm$0.69    & 90.08$\pm$0.34    & 90.02$\pm$0.32    & 91.98$\pm$0.38    & 91.86$\pm$0.37\\
        \rowcolor{orange!20}
         MAGNN-w  & 57.02$\pm$1.23   & 57.36$\pm$0.86    & 59.44$\pm$1.06    & 59.76$\pm$0.93    & 90.24$\pm$0.49    & 90.65$\pm$0.63 & 90.32$\pm$0.77    & 91.32$\pm$0.82   & 88.95$\pm$0.18    & 89.26$\pm$0.20    & 91.23$\pm$0.16    & 91.33$\pm$0.25\\
         \rowcolor{orange!20}
        SHEGNN-w  & 59.56$\pm$0.78   & 61.30$\pm$1.34 & 61.76$\pm$0.62
        & 65.24$\pm$0.73    & 89.32$\pm$0.28    & 89.96$\pm$0.24    & 91.57$\pm$0.21   & 91.71$\pm$0.22    & 91.87$\pm$0.48    & 91.81$\pm$0.36    & 92.45$\pm$0.42    & 92.50$\pm$0.44\\
        \rowcolor{orange!20}
        PSHGCN-w  & 59.68$\pm$0.55   & 61.04$\pm$0.36 & 65.52$\pm$0.52 & 66.03$\pm$0.44   & 89.78$\pm$0.23    & 90.46$\pm$0.25    & 91.58$\pm$0.12    & 91.93$\pm$0.10    & 91.01$\pm$0.26    & 90.97$\pm$0.24    & 92.97$\pm$0.23    & 93.02$\pm$0.19\\
        \hline
        TAPE  & 50.69$\pm$0.30   &   51.06$\pm$0.53   &  53.36$\pm$0.16   &   53.32$\pm$0.35   &   70.56$\pm$0.45   &   70.27$\pm$0.58   &   75.01$\pm$0.44   &   76.15$\pm$0.32   &   78.26$\pm$0.95   &   78.63$\pm$0.97   &   88.91$\pm$0.76   &   88.81$\pm$0.62   \\
        OFA  & 21.50$\pm$0.05 & 21.13 $\pm$0.05    & 22.73$\pm$0.05   & 22.6$\pm$0.05  & 20.80$\pm$0.10        & 20.79$\pm$0.12 & 30.52$\pm$0.08        & 29.89$\pm$0.12&  72.63$\pm$0.23 &72.34$\pm$0.16 &80.32$\pm$0.24   &80.65$\pm$0.28\\
        GOFA & 32.12$\pm$0.20  &  32.29$\pm$0.16  &  33.75$\pm$0.51  &  33.82$\pm$0.26  &  35.90$\pm$0.36  & 35.43$\pm$0.38  &  44.03$\pm$0.50  &  44.52$\pm$0.41  &  78.91$\pm$0.56  & 78.92$\pm$0.73  &  84.28$\pm$0.33  &  84.21$\pm$0.79 \\
        \hline
        \textbf{GHGRL} & \textbf{69.73$\pm$0.53} & \textbf{70.11 $\pm$0.57}    & \textbf{72.13$\pm$0.64}   & \textbf{72.46$\pm$0.62}  & {89.85$\pm$0.23}        & {90.48$\pm$0.18} &  {91.67$\pm$0.34} &{92.17$\pm$0.32} &\textbf{92.71$\pm$0.36}    &\textbf{92.69$\pm$0.30}  &\textbf{94.21$\pm$0.44}&\textbf{94.63$\pm$0.42}\\
                        
        \hline
    \end{tabular}
    \label{tab:main_experiments}
    \vskip -0.1in
\end{table*}

\begin{table*}[ht]\tiny
    \renewcommand{\arraystretch}{1.1}
    \setlength{\tabcolsep}{3.9pt}
    \centering
        \caption{Comparative experiment results for heterogeneous graph datasets with extra diversity. \textbf{Bold} denotes the best performance, \underline{underline} denotes the second best. }
        \vskip -0.1in
    \begin{tabular}{l|cccccc|cccccc}
        \hline
               Datasets & \multicolumn{2}{c}{IMDB-RIR (r=20\%)} & \multicolumn{2}{c}{IMDB-RIR (r=60\%)} & \multicolumn{2}{c|}{IMDB-RIR (r=100\%)}  & \multicolumn{2}{c}{DBLP-RID (r=20\%)} & \multicolumn{2}{c}{DBLP-RID (r=60\%)} & \multicolumn{2}{c}{DBLP-RID (r=100\%)}  \\
        \cline{1-13}
         Metrics & Macro-F1 & Micro-F1 & Macro-F1 & Micro-F1 & Macro-F1 & Micro-F1  & Macro-F1 & Micro-F1  & Macro-F1 & Micro-F1   & Macro-F1 & Micro-F1 \\
        \hline
        Llama3  & 40.25$\pm$0.95  &  39.35$\pm$0.90  &  39.52$\pm$0.38  &  39.67$\pm$0.68  &  39.51$\pm$0.81  &  39.77$\pm$0.51  &  36.56$\pm$1.15  &  46.16$\pm$0.95  &  44.01$\pm$0.85  & 44.89$\pm$0.92  &  43.51$\pm$1.39  &  43.73$\pm$0.56  \\
        \hline   
        TAPE  & \underline{48.59$\pm$1.10}  &  \underline{48.71$\pm$0.61}  &  \underline{45.29$\pm$0.59}  &  \underline{45.01$\pm$0.79}  &  \underline{40.88$\pm$0.74}  &  \underline{40.57$\pm$0.62}  &  \underline{54.87$\pm$0.84}  &  \underline{54.71$\pm$1.11}  &  \underline{52.84$\pm$0.50}  &  \underline{52.14$\pm$0.85}  &  \underline{50.24$\pm$0.51}  &  \underline{50.65$\pm$0.66}  \\

        OFA  & 20.44$\pm$0.12 & 20.78 $\pm$0.26    & 20.84$\pm$0.08   & 20.88$\pm$0.23  & 21.13$\pm$0.12        & 21.21$\pm$0.12 & 31.35$\pm$0.08 & 31.83$\pm$0.17 & 30.33$\pm$0.32 & 30.58$\pm$0.34& 30.52$\pm$0.25   & 30.23$\pm$0.27 \\
        GOFA  & 33.18$\pm$0.62  & 33.91$\pm$0.81  & 31.57$\pm$1.17  & 31.75$\pm$0.84  & 29.16$\pm$0.79  & 29.09$\pm$0.67  & 40.75$\pm$0.58  & 40.28$\pm$0.89  & 38.54$\pm$0.82  & 38.65$\pm$1.00  & 37.11$\pm$0.87  & 37.32$\pm$0.78  \\
        \hline
        \textbf{GHGRL} & \textbf{75.15$\pm$0.43} & \textbf{75.35 $\pm$0.77}    & \textbf{74.72$\pm$0.45}  & \textbf{75.00$\pm$0.42} & \textbf{74.53$\pm$0.56} & \textbf{74.83$\pm$0.52}
        & \textbf{93.47$\pm$0.21}  & \textbf{93.72$\pm$0.25}        & \textbf{92.24$\pm$0.36} &  \textbf{92.78$\pm$0.32} &\textbf{91.20$\pm$0.78} &\textbf{91.83$\pm$0.92} \\ 
        \hline
    \end{tabular}
    \label{tab:ex_experiments}
    \vskip -0.1in
\end{table*}

\begin{table*}[ht]\tiny
    \renewcommand{\arraystretch}{1.1}
    \setlength{\tabcolsep}{5pt}
    \centering
    \caption{Comparative experiment results for HGNNs attached with LLM modules (marked with ``+ LLM''). \textbf{Bold} denotes the best performance, \underline{underline} denotes the second best. }
    \vskip -0.1in
    \begin{tabular}{l|cccccccccc}
        \hline
        Datasets & \multicolumn{2}{c}{IMDB-RIR (r=20\%)} & \multicolumn{2}{c}{IMDB-RIR (r=40\%)} & \multicolumn{2}{c}{IMDB-RIR (r=60\%)} & \multicolumn{2}{c}{IMDB-RIR (r=80\%)} & \multicolumn{2}{c}{IMDB-RIR (r=100\%)} \\
        \cline{2-11}
        Metrics & Macro-F1 & Micro-F1 & Macro-F1 & Micro-F1 & Macro-F1 & Micro-F1 & Macro-F1 & Micro-F1 & Macro-F1 & Micro-F1 \\
        \hline
        GCN + LLM   & 64.26$\pm$0.14 & 64.16$\pm$0.29 & 62.20$\pm$0.18 & 62.86$\pm$0.33 & 59.18$\pm$0.63 & 60.98$\pm$0.71 & 60.05$\pm$0.41 & 60.60$\pm$0.15 & 58.41$\pm$0.35 & 58.59$\pm$0.71 \\
        GAT + LLM   & 65.28$\pm$0.60 & 65.30$\pm$0.70 & 65.78$\pm$0.37 & 65.65$\pm$0.48 & 65.84$\pm$0.76 & 65.77$\pm$0.81 & 65.45$\pm$0.35 & 65.54$\pm$0.75 & 65.25$\pm$0.57 & 65.42$\pm$0.28 \\
        \hline
        HAN + LLM   & 64.89$\pm$0.83 & 65.07$\pm$0.23 & 64.40$\pm$0.83 & 64.49$\pm$0.22 & 64.59$\pm$0.63 & 64.60$\pm$0.55 & 64.00$\pm$0.94 & 64.02$\pm$0.71 & 63.81$\pm$0.46 & 63.90$\pm$0.30 \\
        MAGNN + LLM & 61.88$\pm$0.56 & 61.92$\pm$0.47 & 61.25$\pm$0.32 & 61.34$\pm$0.18 & 61.42$\pm$0.46 & 61.36$\pm$0.87 & 61.41$\pm$0.35 & 61.39$\pm$0.24 & 61.34$\pm$0.19 & 61.32$\pm$0.52 \\
        SeHGNN + LLM & 68.35$\pm$0.50 & 68.82$\pm$0.34 & 68.74$\pm$0.61 & 68.52$\pm$0.47 & 68.21$\pm$0.81 & 68.62$\pm$0.64 & 67.98$\pm$0.30 & 68.24$\pm$0.39 & 67.76$\pm$0.44 & 68.06$\pm$0.33 \\
        PSHGCN + LLM & \underline{71.83$\pm$0.23} & \underline{72.18$\pm$0.47} & \underline{72.30$\pm$0.91} & \underline{72.18$\pm$0.61} & \underline{72.36$\pm$0.78} & \underline{72.77$\pm$0.91} & \underline{72.50$\pm$0.21} & \underline{72.88$\pm$0.73} & \underline{72.28$\pm$0.60} & \underline{72.65$\pm$0.14} \\
        \hline
         \rowcolor{orange!20}
        HAN-w + LLM & 63.41$\pm$0.90 & 63.52$\pm$0.38 & 63.28$\pm$0.28 & 63.56$\pm$0.55 & 63.29$\pm$0.89 & 63.30$\pm$0.08 & 63.26$\pm$0.59 & 63.45$\pm$0.57 & 63.88$\pm$0.63 & 63.01$\pm$0.24 \\
         \rowcolor{orange!20}
        MAGNN-w + LLM & 61.84$\pm$0.32 & 61.87$\pm$0.16 & 61.16$\pm$0.72 & 61.24$\pm$0.41 & 60.58$\pm$0.51 & 60.71$\pm$0.86 & 60.94$\pm$0.38 & 61.12$\pm$0.19 & 61.24$\pm$0.11 & 61.33$\pm$0.68 \\
         \rowcolor{orange!20}
        SeHGNN-w + LLM & 66.24$\pm$0.57 & 66.33$\pm$0.37 & 66.12$\pm$0.58 & 66.18$\pm$0.55 & 66.02$\pm$0.26 & 66.09$\pm$0.35 & 65.82$\pm$0.88 & 65.96$\pm$0.96 & 65.79$\pm$0.39 & 65.92$\pm$0.71 \\
         \rowcolor{orange!20}
        PSHGCN-w + LLM & 71.43$\pm$0.32 & 71.53$\pm$1.12 & 71.06$\pm$0.63 & 71.32$\pm$0.66 & 70.89$\pm$0.38 & 71.07$\pm$0.21 & 70.55$\pm$0.25 & 70.86$\pm$0.22 & 70.21$\pm$1.06 & 70.43$\pm$0.83 \\
        \hline
        \textbf{GHGRL} & \textbf{75.15$\pm$0.43} & \textbf{75.35$\pm$0.77} & \textbf{74.48$\pm$0.51} & \textbf{74.90$\pm$0.63} & \textbf{74.72$\pm$0.45} & \textbf{75.00$\pm$0.42} & \textbf{74.53$\pm$0.56} & \textbf{74.83$\pm$0.52} & \textbf{74.93$\pm$0.46} & \textbf{75.15$\pm$0.51} \\
        \hline
    \end{tabular}
    \label{tab:ririmdb_experiments}
    \vskip -0.1in
\end{table*}

\begin{table}[ht]\tiny
    \renewcommand{\arraystretch}{1.1}
    \setlength{\tabcolsep}{5pt}
    \centering
    \caption{Comparative experimental results for the homogeneous dataset. \textbf{Bold} indicates the best performance, while \underline{underline} indicates the second best.}
    \vskip -0.1in
    \begin{tabular}{l|cc}
        \hline
        Dataset & \multicolumn{2}{c}{Wiki-CS} \\
        \cline{2-3}
        Metrics & Macro-F1 & Micro-F1 \\
        \hline
        GCN   & 69.78$\pm$0.53 & 75.10$\pm$0.58 \\
        GAT   & 70.88$\pm$0.50 & 78.04$\pm$0.63 \\
        \hline
        TAPE  & 77.30$\pm$0.59  & 77.24$\pm$0.67 \\
        OFA & 77.69$\pm$0.12 & 78.32$\pm$0.15 \\
        GOFA & 78.65$\pm$0.68  & 78.74$\pm$0.95 \\
        \hline
        \textbf{GHGRL} & \textbf{80.69$\pm$0.60} & \textbf{81.39$\pm$0.27} \\
        \hline
    \end{tabular}
    \label{tab:homogeneous_experiments}
    \vskip -0.1in
\end{table}


\subsection{Comparison with State of the Art Methods}
\subsubsection{Baselines.}
For baseline methods, we compared our approach with three categories of baselines: 1) general GNN backbone networks, including GCN \cite{DBLP:conf/iclr/KipfW17} and GAT  \cite{DBLP:conf/iclr/VelickovicCCRLB18}, 2) HGNN methods, including HAN \cite{DBLP:conf/www/WangJSWYCY19}, MAGNN \cite{DBLP:conf/www/0004ZMK20}, SeHGNN \cite{yang2023simple} and PSHGCN \cite{DBLP:conf/www/HeWFHLSY24} and 3) more generalized graph representation learning methods that combines GNN and LLM, including TAPE \cite{DBLP:conf/iclr/HeB0PLH24}, OFA \cite{liuone}, and GOFA \cite{kong2024gofagenerativeoneforallmodel}. 

\subsubsection{Datasets.}
We utilized existing commonly used heterogeneous and homogeneous graph representation learning datasets, as well as more challenging heterogeneous graph datasets that we newly constructed. Specifically, we employed the IMDB, DBLP, ACM \cite{zhang2019heterogeneous} and Wiki-CS \cite{DBLP:journals/corr/abs-2007-02901} datasets , and we reported the test accuracy under varying amounts of training data. Additionally, we constructed two new datasets, the \textit{\textbf{R}andom \textbf{I}nformation \textbf{R}eplacement on \textbf{IMDB}} (IMDB-RIR) and the \textit{\textbf{R}andom \textbf{I}nformation \textbf{D}eletion on \textbf{DBLP}} (DBLP-RID):
We utilized both commonly used heterogeneous graph representation learning datasets and more challenging datasets that we newly constructed. Specifically, we employed the IMDB, DBLP, and ACM datasets \cite{zhang2019heterogeneous}, and reported the test accuracy with varying amounts of training data. Additionally, we constructed two new datasets: \textit{\textbf{IMDB} dataset with \textbf{R}andom \textbf{I}nformation \textbf{R}eplacement} (IMDB-RIR) and \textit{\textbf{DBLP} dataset with \textbf{R}andom \textbf{I}nformation \textbf{D}eletion} (DBLP-RID).

\begin{itemize}
\item \textbf{IMDB-RIR.} Based on the IMDB dataset, we performed searches on Google using the textual information of the nodes in the IMDB dataset. We then saved the top 10 search results for each node. Subsequently, we randomly selected results from these top 10 and used them to replace the node attributes in the IMDB dataset. As a result, the constructed dataset contains information in various uncertain formats, thereby increasing the complexity of the tasks.

\item \textbf{DBLP-RID.} Based on the DBLP dataset, we randomly deleted portions of the node textual information, creating a new graph dataset with partially missing node information.
\end{itemize}

These datasets introduce further diversity into heterogeneous datasets and are utilized for extra comparison. Further details can be found in \textbf{Appendix} D.1.

\subsubsection{Settings.}
We followed the basic settings outlined in OFA \cite{liuone} and used Llama 3 \cite{dubey2024llama3herdmodels} as the LLM for both our method and the baseline methods to ensure a fair comparison. Additionally, we adjusted the proportion of training data in the datasets to compare test results under different conditions. For all experimental results, we conducted five independent runs and reported the mean ± standard deviation.
The specific experimental setup, including hyperparameters and the environment used, is detailed in \textbf{Appendix} D. 

\subsubsection{Results on heterogeneous graph datasets.}
Table \ref{tab:main_experiments} demonstrates the results of experiments on IMDB, DBLP, and ACM. Since our approach does not use the node type or edge type information included in the heterogeneous graph datasets as input, for better analysis, we compared our method with HGNN baselines that both use and do not use this information. We mark the methods that do not utilize this information with ``-w''. In the results, we can see that our method achieves either the best performance or performance comparable to methods that use additional type information on all datasets, demonstrating the capability of GHGRL. 

\subsubsection{Results on heterogeneous graph datasets with extra diversity.}
Table \ref{tab:ririmdb_experiments} demonstrates the results of experiments on IMDB-RIR and DBLP-RID. We denote $r$ as the proportion of newly constructed data used in the dataset, e.g., $20\%$ denotes we utilize $20\%$ of the total amount of newly constructed data. Since within IMDB-RIR and DBLP-RID, the node features have been modified by the additional information we introduced, they no longer adhere to a standard format. Consequently, GNN and HGNN-based methods can no longer process this information without additional help, so we did not include comparisons with these methods. Additionally, we used our LLM, Llama 3, to directly classify the nodes. Here, we can see that while LLM-based methods can somewhat handle our newly constructed dataset, GHGRL still achieved the best performance, significantly surpassing other baseline methods and demonstrating its capability.

Furthermore, in Table \ref{tab:ex_experiments}, we also integrated the LLM processing module we used into other HGNN methods to output unified features for further comparison on the IMDB-RIR dataset. As shown in the results, even under these conditions, GHGRL still achieved the best performance, significantly outperforming other methods. This demonstrates the strong compatibility between our PAGNN and the LLM module.

\subsubsection{Results on homogeneous graph dataset.}
We also conducted method comparisons on homogeneous graph datasets. The results show that GHGRL can achieve better performance on homogeneous graphs as well, indicating that its mechanism positively enhances graph representation learning even on standard homogeneous graph data. Further experimental results can be found in \textbf{Appendix} E.

\subsection{In-Depth Analysis}

\begin{figure}[ht]
    \centering
    \subfigure[Input.]{%
        \includegraphics[width=0.11\textwidth]{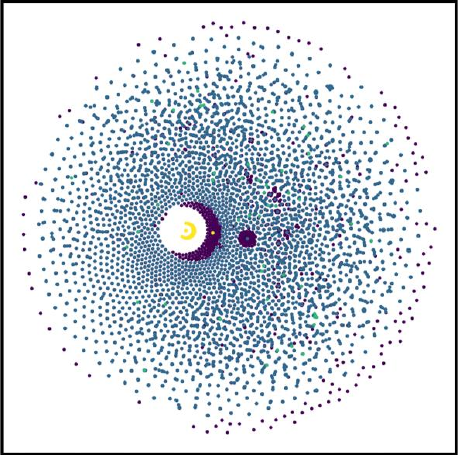}
        \label{fig:subfig1}
    }
    \hspace{1mm}
    \subfigure[LLM Processed.]{%
        \includegraphics[width=0.11\textwidth]{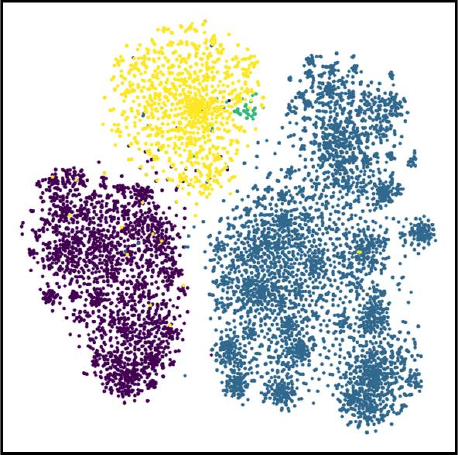}
        \label{fig:subfig2}
    }
    \hspace{1mm}
    \subfigure[Output.]{%
        \includegraphics[width=0.11\textwidth]{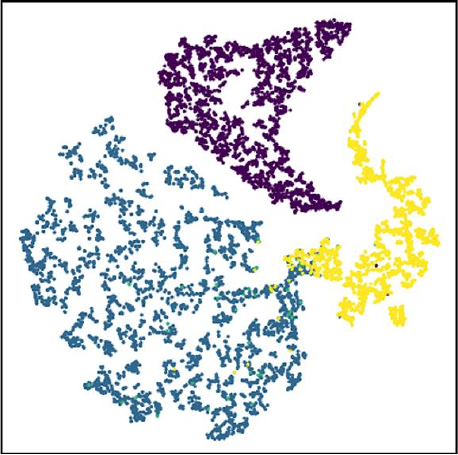}
        \label{fig:subfig3}
    }
    \vskip -0.1in

    \caption{Data representations at different stages of the model after dimensionality reduction using the t-SNE method. Different colors represent distinct types of nodes.}
    \vskip -0.1in
    \label{fig:tsne1}
\end{figure}

\begin{figure}[ht]
    \centering
    \subfigure[Input.]{%
        \includegraphics[width=0.11\textwidth]{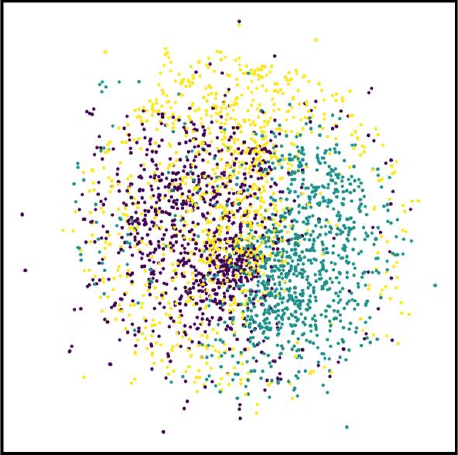}
        \label{fig:subfig4}
    }
    \hspace{1mm}
    \subfigure[LLM Processed.]{%
        \includegraphics[width=0.11\textwidth]{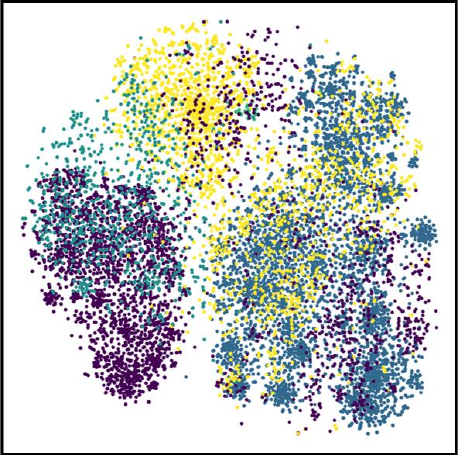}
        \label{fig:subfig5}
    }
    \hspace{1mm}
    \subfigure[Output.]{%
        \includegraphics[width=0.11\textwidth]{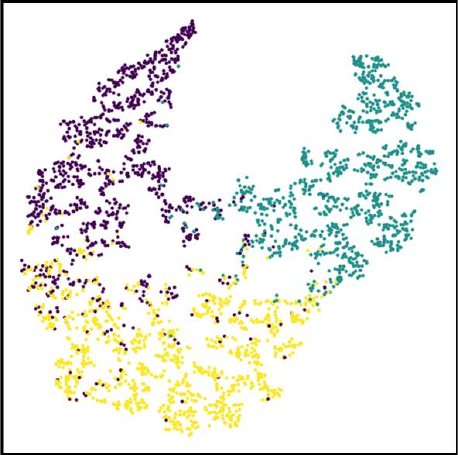}
        \label{fig:subfig6}
    }
    \vskip -0.1in
    \caption{Data representations at different stages of the model after dimensionality reduction using the t-SNE method. Different colors represent distinct classes of nodes.}
    \vskip -0.1in
    \label{fig:tsne2}
\end{figure}

\begin{figure}[ht]
    \centering
    \subfigure[IMDB.]{%
        \includegraphics[width=0.4\textwidth]{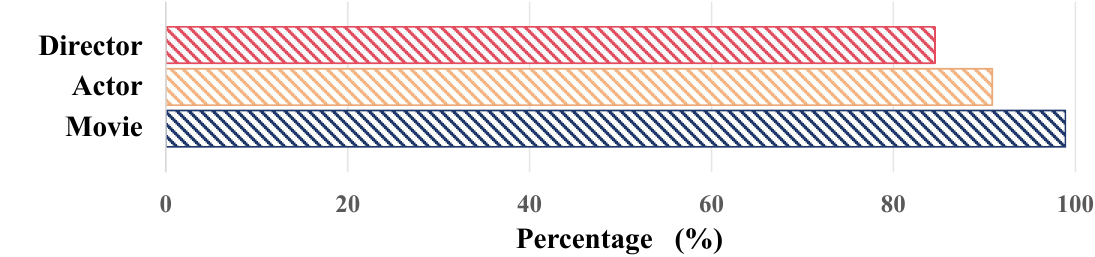}
        \label{fig:barsubfig1}
    }
    \vskip -0.1in
    \subfigure[DBLP.]{%
        \includegraphics[width=0.4\textwidth]{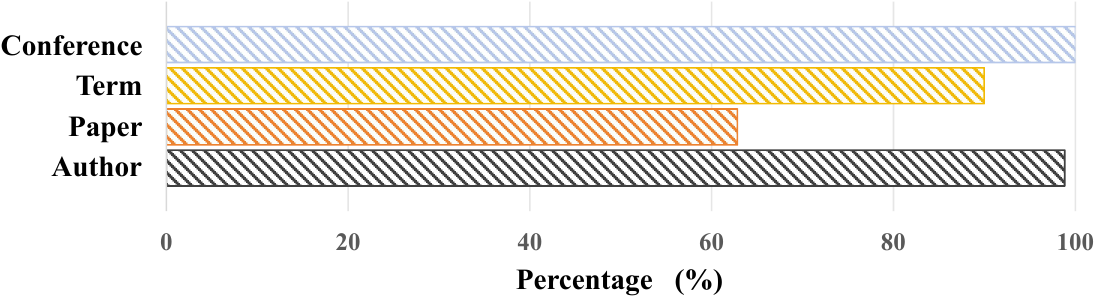}
        \label{fig:barsubfig2}
    }
    \vskip -0.1in
    \subfigure[ACM.]{%
        \includegraphics[width=0.4\textwidth]{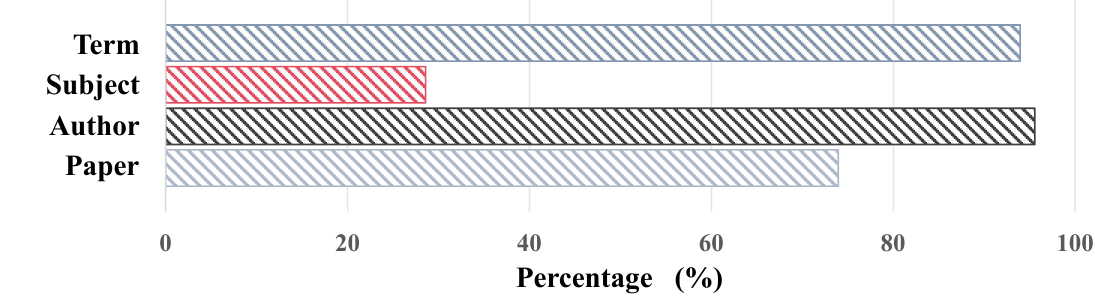}
        \label{fig:barsubfig3}
    }
    \vskip -0.1in
    \caption{Demonstration of different methods.}
    \vskip -0.15in
    \label{fig:barchart}
\end{figure}


\subsubsection{Feature visualization.}
We visualized the node features of the model at different stages on the ACM dataset using the t-SNE method, as shown in Figures \ref{fig:tsne1} and \ref{fig:tsne2}. From these figures, it is evident that at the input stage, the node features are highly mixed. However, after being processed by the LLM, these features display multiple dispersed clusters. This indicates that the LLM has leveraged its knowledge to perform a more detailed grouping of the samples. However, this grouping does not align with the desired three-class categorization of the nodes, as some node features remain intermixed. Finally, after processing by PAGNN, our model successfully categorizes the nodes into three distinct groups according to their classes, demonstrating that PAGNN has further refined the information extracted from the LLM's output, ultimately leading to a more optimal result.

\subsubsection{LLM processing analysis.}
Here, we report the statistics on the match between the node types estimated by the model and the actual node types in the IMDB, DBLP, and ACM datasets. The proportion of correctly classified types for each category is summarized in Figure \ref{fig:barchart}. It is evident that our model does not accurately estimate all types. This reveals an interesting phenomenon: our LLM Processing module classifies nodes in the dataset based on its own internal knowledge. Moreover, the results of the aforementioned comparative experiments demonstrate that our model outperforms other models, indicating that GHGRL effectively leverages PAGNN to adapt to the estimations made by the LLM Processing module. As a result, even when there is a discrepancy between the classification and the actual dataset, our model can still achieve satisfactory performance.

\section{Conclusion}
In this paper, we propose an innovative approach called GHGRL, which integrates LLM and GNN using an adaptive parameter selection method. This approach enhances the generalization capability for handling heterogeneous graph data, offering a new perspective for processing more complex and irregularly structured graph data.

\section*{Acknowledgment}
We would like to express our sincere gratitude to the reviewers of this paper, as well as the Program Committee and Area Chairs, for their valuable comments and suggestions. This work is supported by the CAS Project for Young Scientists in Basic Research, Grant No. YSBR-040. 

\bibliography{aaai25}

\newpage
\appendix
\onecolumn
\section{A. Extended Related Works}
\subsection{A.1. Heterogenous Graph Representation Learning.} 
The methods for heterogeneous graph representation learning can be categorized into metapath-based and metapath-free approaches. Among them, metapath-based methods leverage heterogeneous graph neural networks to first aggregate features of neighbors with the same semantics and then integrate different semantics \cite{DBLP:journals/air/BingYZMMQ23, DBLP:conf/nips/YunJKKK19, zhang2019heterogeneous, DBLP:conf/www/WangJSWYCY19, DBLP:conf/www/0004ZMK20,yang2023simple}. HetGNN \cite{zhang2019heterogeneous} uses random walks to integrate neighbors at various distances and aggregates data based on node types. HAN \cite{DBLP:conf/www/WangJSWYCY19} differentiates between different semantics using metapaths and propagates data accordingly. MAGNN \cite{DBLP:conf/www/0004ZMK20} incorporates all nodes in the metapath during data propagation, rather than solely utilizing endpoints. SeHGNN \cite{yang2023simple} uses a single-layer structure with long metapaths to extend the receptive field and a transformer-based module to fuse features from different metapaths. Metapath-free methods embed semantic information into propagated messages using attention mechanisms and other techniques \cite{zhu2019relation, DBLP:conf/aaai/HongGLYLY20,DBLP:conf/kdd/FanZHSHML19, DBLP:conf/kdd/LvDLCFHZJDT21,DBLP:conf/icml/ZhouSYZ023,DBLP:conf/www/HeWFHLSY24}. RSHN \cite{DBLP:conf/kdd/FanZHSHML19} obtains global embeddings for different edge types and uses a combination of neighbor features and edge type embeddings for feature aggregation at each layer. HGB \cite{DBLP:conf/kdd/LvDLCFHZJDT21} uses a multi-layer GAT network to distinguish heterogeneous nodes and uses both node features and learnable edge-type embeddings to generate attention values. PSHGCN \cite{DBLP:conf/www/HeWFHLSY24} uses positive spectral heterogeneous graph convolution to learn valid heterogeneous graph filters. These methods all require prior knowledge of node types and are typically used on datasets where node types are known. However, this limitation restricts the application of these methods in the broader field of data mining.

\subsection{A.2. LLMs for Graphs.}
With the emergence of various LLM methods, the use of LLMs for graph representation learning is gradually becoming a research hotspot. Relevant studies can be classified into two types. One type enriches node representation based on prompt learning and then completes subsequent graph data processing tasks based on GNN \cite{DBLP:conf/www/0001RTYC24, DBLP:journals/corr/abs-2310-04560, DBLP:journals/sigkdd/ChenMLJWWWYFLT23, DBLP:conf/nips/HuangRCK0LL23, liuone, DBLP:conf/sigir/Tang00SSCY024}. Among these works, TAPE \cite{DBLP:conf/iclr/HeB0PLH24} prompts an LLM for zero-shot classification, extracts its explanations, and uses an interpreter to convert these into features for downstream GNNs. OFA \cite{liuone} unifies different graph data by describing nodes and edges in natural language and uses a language model to encode text attributes, converting them into feature vectors in the same embedding space. It also introduces the concept of attention nodes and a new graph prompt paradigm to solve different tasks without fine-tuning. GOFA \cite{kong2024gofagenerativeoneforallmodel} optimizes OFA by interleaving randomly initialized GNN layers with frozen pre-trained language model LLM, organically combining semantic and structural modeling capabilities. The other type directly converts graph data into text to input into the LLM for processing and directly obtains the answer \cite{DBLP:journals/corr/abs-2403-04483, DBLP:journals/corr/abs-2403-04780, DBLP:journals/corr/abs-2312-10372,DBLP:conf/nips/WangFHTHT23,DBLP:journals/corr/abs-2305-15066,DBLP:conf/kdd/SunCLLG23}. NLGraph \cite{DBLP:conf/nips/WangFHTHT23} proposes two processing methods: Build-a-Graph Prompting and Algorithmic Prompting. GPT4Graph \cite{DBLP:journals/corr/abs-2305-15066} focuses on the LLMs' capabilities in graph understanding. All in One \cite{DBLP:conf/kdd/SunCLLG23} introduces meta-learning to effectively learn better initialization of multi-task prompts for graphs, making the prompt framework more reliable and generalizable for different tasks. However, all the aforementioned methods deal with homogeneous data and cannot effectively handle heterogeneous data. They also require some preprocessing of the input graph data itself, which limits the application of these methods in data mining. In fact, the potential of graph representation learning methods that integrate LLMs should far exceed this. Therefore, we have designed a method to integrate LLM to handle heterogeneous graph attribute information of any format and type, with these formats and attributes not needing to be known in advance. This greatly expands the application scope of graph representation learning methods in data mining.

\section{B. Proofs}

\subsection{B.1. Proof of Theorem \ref{thm:m}}
\paragraph{Theorem 1.}
Given a connected graph $G = \{\mathcal{V}, \mathcal{E}\}$ with node features $\{\bm{x}_{i}\}_{i=1}^{|\mathcal{V}|}$ and LLM $f(\cdot)$, $\widetilde{g}(\cdot)$ can avoid the over-smoothing described in Equation \ref{eq:os} for the node features, i.e., we have:
\begin{equation}
    \lim_{L \to + \infty} \widetilde{g}(\{f(\bm{x}_{j})\}_{j=1}^{|\mathcal{V}|}, \mathcal{E}) = \begin{bmatrix}\Tilde{\bm{h}}_{1}^{(L)}& \Tilde{\bm{h}}_{2}^{(L)}& \cdots& \Tilde{\bm{h}}_{|\mathcal{V}|}^{(L)}\end{bmatrix}^\top,
    \label{eq:dsim2}
\end{equation}
where for node $i$ and $j$ that satisfied $\phi(i) \neq \phi(j)$, $\Tilde{\bm{h}}_{i}$ and $\Tilde{\bm{h}}_{j}$ are linearly independent.

\begin{proof}
According to the theorem, we have the output feature $H$ with $g(\cdot)$ as follows:
\begin{equation}
    H = g\Big(\{f(\bm{x}_{j})\}_{j=1}^{|\mathcal{V}|},\mathcal{E}\Big) = SGC^{(L)} \circ SGC^{(L-1)} \circ \cdots \circ  SGC^{(1)}\Big(\{f(\bm{x}_{j})\}_{j=1}^{|\mathcal{V}|},\mathcal{E}\Big),
    \label{eq:pini}
\end{equation}
where $SGC^{(l)}(\cdot)$ denotes the $l$-th spectral graph convolution computation. We can also denote the $l$-th computation as follows:
\begin{equation}
    H^{(l+1)} = SGC^{(l)}\Big(H^{(l)},\mathcal{E}\Big),
    \label{eq:exp}
\end{equation}
where $H^{(l+1)}$ denote the output node features of the $l$-th layer and $H^{(l)}$ denotes the inputs. Based on the properties of spectral graph convolution \cite{DBLP:conf/iclr/KipfW17}, we can expand Equation \ref{eq:exp} as follows:
\begin{align}
    H^{(l+1)} &= SGC^{(l)}\Big(H^{(l)},\mathcal{E}\Big) \nonumber\\
    &=  (I + D^{-\frac{1}{2}}AD^{-\frac{1}{2}})
    \begin{bmatrix} \bm{h}_{1}^{(l)} & \bm{h}_{2}^{(l)} & \cdots & \bm{h}_{|\mathcal{V}|}^{(l)} \end{bmatrix}^\top W^{(l)} ,
\end{align}
where $\bm{h}_{j}^{(l)}$ denotes the $j$-th node representation of $l$-th layer, $W^{l}$ is the $l$-th parameter matrix. Next, we can obtain the final output $H$ calculated with $L$ layers of graph convolution as follows:
\begin{align}
    H &=  (I + D^{-\frac{1}{2}}AD^{-\frac{1}{2}})  \Big( (I + D^{-\frac{1}{2}}AD^{-\frac{1}{2}}) \ \cdots \ \nonumber\\ 
    & \  \Big( (I + D^{-\frac{1}{2}}AD^{-\frac{1}{2}})
    \begin{bmatrix} \bm{h}_{1}^{(1)} & \bm{h}_{2}^{(1)} & \cdots & \bm{h}_{|\mathcal{V}|}^{(1)} \end{bmatrix}^\top W^{(1)}\Big) \ \cdots \  W^{(L-1)}\Big)W^{(L)} .
    \nonumber\\ 
    &=    (\widetilde{D}^{-\frac{1}{2}}\widetilde{A}\widetilde{D}^{-\frac{1}{2}})^{L}
    \begin{bmatrix} \bm{h}_{1}^{(1)} & \bm{h}_{2}^{(1)} & \cdots & \bm{h}_{|\mathcal{V}|}^{(1)} \end{bmatrix}^\top W^{\text{sum}},
    \label{eq:hhh}
\end{align}
where $\widetilde{D}$ and $\widetilde{A}$ are the degree matrix and adjacency matrix containing self-loops. Due to the associative property of matrix multiplication, $W^{\text{sum}}$ represents the precomputed product of all parameter matrices. Within Equation \ref{eq:hhh}, $\begin{bmatrix} \bm{h}_{1}^{(1)} & \bm{h}_{2}^{(1)} & \cdots & \bm{h}_{|\mathcal{V}|}^{(1)} \end{bmatrix}$ denotes the init input of the spectral graph convolution, i.e.:
\begin{equation}
    \begin{bmatrix} \bm{h}_{1}^{(1)} & \bm{h}_{2}^{(1)} & \cdots & \bm{h}_{|\mathcal{V}|}^{(1)} \end{bmatrix} =     \begin{bmatrix} f(\bm{x}_{1}) & f(\bm{x}_{2}) & \cdots & f(\bm{x}_{|\mathcal{V}|}) \end{bmatrix}.
\end{equation}
As $\widetilde{D}^{-\frac{1}{2}}\widetilde{A}\widetilde{D}^{-\frac{1}{2}}$ is a real symmetric matrix, real symmetric matrices can always be orthogonally diagonalized, and therefore can always undergo standard orthogonal decomposition. Therefore we have:
\begin{align}
    H &= (P \Lambda P^{\top})^L \begin{bmatrix} \bm{h}_{1}^{(1)} & \bm{h}_{2}^{(1)} & \cdots & \bm{h}_{|\mathcal{V}|}^{(1)} \end{bmatrix}^\top W^{\text{sum}},
\end{align}
where $\Lambda$ is the eigenvalue matrix, and $P$ is the orthonormal eigenvector matrix. Then we have:
\begin{align}
    H &= P \Lambda^L P^{\top} \begin{bmatrix} \bm{h}_{1}^{(1)} & \bm{h}_{2}^{(1)} & \cdots & \bm{h}_{|\mathcal{V}|}^{(1)} \end{bmatrix}^\top W^{\text{sum}}.
\end{align}
According to \cite{chung1997spectral}, the values of $\Lambda^L$ will fall into $(-1,1]$. Therefore, after repeatedly multiplying $ P^{\top} \begin{bmatrix} \bm{h}_{1}^{(1)} & \bm{h}_{2}^{(1)} & \cdots & \bm{h}_{|\mathcal{V}|}^{(1)} \end{bmatrix}$ with $\Lambda$ from the left, the result will be an eigenvalue matrix with values of 0 or 1. Based on \cite{DBLP:conf/aaai/LiHW18}, and given that $G$ is a connected graph, the eigenvectors corresponding to the eigenvalues in $\Lambda^L$ are all unit vectors. Then, when $L \rightarrow + \infty$, we have:
\begin{align}
    H &= \begin{bmatrix}\theta_{1} \bar{\bm{h}}'& \theta_{2}\bar{\bm{h}}'& \cdots& \theta_{|\mathcal{V}|}\bar{\bm{h}}'\end{bmatrix}^\top W^{\text{sum}},
    \label{eq:hsame}
\end{align}
where $\bar{\bm{h}}' \in \mathbb{R}^{D}$. It can be deviate form Equation \ref{eq:hsame} that:
\begin{align}
    H &= \begin{bmatrix}\theta_{1} \bar{\bm{h}}& \theta_{2}\bar{\bm{h}}& \cdots& \theta_{|\mathcal{V}|}\bar{\bm{h}}\end{bmatrix}^\top,
\end{align}
where $\bar{\bm{h}}$ is certain vector and $\bar{\bm{h}} \in \mathbb{R}^{D}$. According to the theorem, $\widetilde{g}(\cdot)$ attaches the simplified GHGRL module, therefore, we have the new output feature $\widetilde{H}$ as:
\begin{align}
    \widetilde{H} &= \widetilde{g}(\{f(\bm{x}_{j})\}_{j=1}^{|\mathcal{V}|}, \mathcal{E}),
    \nonumber\\
    &=  \underbrace{(P \Lambda P^{\top})\text{Selection}\Big((P \Lambda P^{\top}) \text{Selection}\Big( \ \cdots  \ (P \Lambda P^{\top})\text{Selection} \Big( \ }_{L \ \text{times}}
      \nonumber\\ 
    & \ \begin{bmatrix} \widetilde{\bm{h}}_{1}^{(1)} & \widetilde{\bm{h}}_{2}^{(1)} & \cdots & \widetilde{\bm{h}}_{|\mathcal{V}|}^{(1)} \end{bmatrix}^\top \underbrace{\Big) \ \cdots \  \Big)\Big)}_{L \ \text{times}},
    \nonumber\\ 
\end{align}
where $\text{Selection}(\cdot)$ denotes the parameter selection operation of $\widetilde{g}(\cdot)$, which no longer satisfies the associative property. Furthermore, we have:
\begin{align}
    \widetilde{H}^{(l+1)} 
    &=  (P \Lambda P^{\top}) \text{Selection}\Big(
    \begin{bmatrix} \widetilde{\bm{h}}_{1}^{(l)} & \widetilde{\bm{h}}_{2}^{(l)} & \cdots & \widetilde{\bm{h}}_{|\mathcal{V}|}^{(l)} \end{bmatrix}^\top \Big)
    \nonumber\\ 
    &=  (P \Lambda P^{\top}) \Big(\begin{bmatrix} \widetilde{\bm{h}}_{1}^{(l)}W^{(l)}_{1} + \bm{b}_{1}^{(l)} & \widetilde{\bm{h}}_{2}^{(l)}W^{(l)}_{2} + \bm{b}_{2}^{(l)} & \cdots & \widetilde{\bm{h}}_{|\mathcal{V}|}^{(l)}W^{(l)}_{|\mathcal{V}|} + \bm{b}_{|\mathcal{V}|}^{(l)} \end{bmatrix}^\top \Big),
\end{align}
where $W_{i}^{(l)} = W_{j}^{(l)}$ and $\bm{b}_{i}^{(l)} = \bm{b}_{j}^{(l)}$ for node $i$ and $j$ that belongs to the same type, while $W_{i}^{(l)} \neq W_{j}^{(l)}$ and $\bm{b}_{i}^{(l)} \neq \bm{b}_{j}^{(l)}$ for otherwise. If $\bm{h}_{i}^{(l)} = \theta \bm{h}_{j}^{(l)}$ and node $i$ and $j$ that belongs to different type, we still have $\bm{h}_{i}^{(l)}W_{i}^{(l)}+ \bm{b}_{i}^{(l)} \neq \theta \bm{h}_{j}^{(l)}W_{j}^{(l)}+ \bm{b}_{j}^{(l)}, \forall \theta \in \mathbb{R}$. Therefore, we can conclude that after infinite layers, Equation \ref{eq:dsim2} still holds, and the theorem is proved.
\end{proof}

\subsection{B.2. Proof of Corollary \ref{thm:cly}}

\paragraph{Corollary 2.}
For conditions given in Theorem \ref{thm:m}, if node $i$ and $j$ satisfied $\phi(i) = \phi(j)$, if $i$ and $j$ do not share same set of neighbours, then $\Tilde{\bm{h}}^{(L)}_{i}$ and $\Tilde{\bm{h}}^{(L)}_{j}$ are not necessarily linear dependent.

\begin{proof}
    We demonstrate corollary \ref{thm:cly} through expanding Equation \ref{eq:shgnnex}. Formally, we have:
\begin{align}
            \bm{h}_{v}^{(l+1)} &= \bm{h}_{v}^{(l)} + AGG\left(\bm{h}_{v}^{(l)}W^{ [\phi(v)] } + B^{ [\phi(v)] } ,u \in \mathcal{N}\left(v\right)\right)
            \nonumber\\
            &=  \bm{h}_{v}^{(l)} + \Bigg(\left(\bm{h}_{u_{1}}^{(l)}W^{ [\phi(u_{1})] } + B^{ [\phi(u_{1})] }\right)  + \left(\bm{h}_{u_{2}}^{(l)}W^{ [\phi(u_{2})] } + B^{ [\phi(u_{2})] }\right) +  \ldots +
            \nonumber\\
            & \quad \left(\bm{h}_{u_{|\mathcal{N}(v)|}}^{(l)}W^{ [\phi(u_{|\mathcal{N}(v)|})] } + B^{ [\phi(u_{|\mathcal{N}(v)|})] }\right) \Bigg).
\end{align}
Then, for node $i$ and $j$ that satisfied $\phi(i) = \phi(j)$, we have:
\begin{align}
            \bm{h}_{i}^{(l+1)} =  \bm{h}_{i}^{(l)} + \sum_{u}^{|\mathcal{N}(i)|} \left(\bm{h}_{u}^{(l)}W^{ [\phi(u)] } + B^{ [\phi(u)] }\right), u \in \mathcal{N}\left(i\right),
\end{align}
and:
\begin{align}
            \bm{h}_{j}^{(l+1)} =  \bm{h}_{j}^{(l)} + \sum_{u}^{|\mathcal{N}(j)|} \left(\bm{h}_{u}^{(l)}W^{ [\phi(u)] } + B^{ [\phi(u)] }\right), u \in \mathcal{N}\left(v\right),
\end{align}
where even when $\bm{h}_{j}^{(l)}$ and $\bm{h}_{i}^{(l)}$ is linearly dependent if the adjacent nodes of these two nodes are not all linearly dependent, there exist a set of parameters that let $\bm{h}_{j}^{(l+1)}$ and $\bm{h}_{i}^{(l+1)}$ be linearly independent. The corollary is proved.

\end{proof}

\section{C. Implementation Details }
\label{apx:id}
In this section, we provide a further introduction and practical demonstration of the prompts used. Specifically, we adopt the following prompt for type generation.

\begin{tcolorbox}[colframe=black, colback=gray!20, coltitle=black, coltext=black, breakable,  boxrule=0.5mm, title=\textcolor{white}{Type Generation Prompt}]
\textbf{Given data:}

The following contents are the descriptions of nodes within a graph: \textbf{\textcolor{blue}{$\textlangle\text{node attribute 1}\textrangle$}}; \textbf{\textcolor{blue}{$\textlangle\text{node attribute 2}\textrangle$}}; \textbf{\textcolor{blue}{$\textlangle\text{node attribute 3}\textrangle$}}; \textcolor{blue}{......}; \textbf{\textcolor{blue}{$\textlangle\text{node attribute n}\!>$}}.

\textbf{Answer the following questions:}

\begin{enumerate}
    \item Which \textbf{\textcolor{orange}{$\textlangle\text{format type number}\textrangle$}} types can these nodes be divided according to their format? Provide the names of these types and separate them with semicolons.
    
    \item Which \textbf{\textcolor{red}{$\textlangle\text{content type number}\textrangle$}} types can these nodes be divided according to their content? Provide the names of these types and separate them with semicolons.
\end{enumerate}
\textbf{Please only provide answers and separators strictly in the given order.}
\end{tcolorbox}

$\textbf{\textcolor{black}{$\textlangle\text{node attribute i}\textrangle$}}$ denotes the content of the $i$-th node attribute, $\textbf{\textcolor{black}{$\textlangle\text{format type number}\textrangle$}}$ and $\textbf{\textcolor{black}{$\textlangle\text{content type number}\textrangle$}}$ denotes the numbers of how many types to be generated. This prompt guides the model to output possible node types in a fixed format, which will be used to inform subsequent experiments based on those types. 

Furthermore, we conduct analysis upon each node $v$'s feature $\bm{x}_{v}$ with the following prompt. 

\begin{tcolorbox}[colframe=black, colback=gray!20, coltitle=black, coltext=black,
  title=\textcolor{white}{LLM Processing Prompt}]
\textbf{Given data:} 

The following content is the descriptions of a node within a graph: \textcolor{blue}{\textbf{$\textlangle$node attribute$\textrangle$}}.

\textbf{Answer the following questions:}

\begin{enumerate}
    \item Provide a description of \textcolor{blue}{\textbf{$\textlangle$node attribute$\textrangle$}}. The description should be as comprehensive and detailed as possible.
    \item Which format type within \textcolor{orange}{\textbf{$\textlangle$format type set$\textrangle$}} does the node belong to? Provide the name of the type.
    \item Provide the reason for the answer of Question 2.
    \item Regarding Question 2, how certain are you of your answer? Provide a confidence score between 0 and 1.
    \item Which content type within \textcolor{red}{\textbf{$\textlangle$content type set$\textrangle$}} does the node belong to? Provide the name of the type.
    \item Regarding Question 4, how certain are you of your answer? Provide a confidence score between 0 and 1.
\end{enumerate}
\textbf{Please only provide answers and separators strictly in the given order.}
\end{tcolorbox}

Within the prompt, \textcolor{black}{\textbf{$\textlangle$node attribute$\textrangle$}} denotes the node attribute information,  \textcolor{black}{\textbf{$\textlangle$format type set$\textrangle$}} denotes set $\Phi^{\text{fmt}}$, \textcolor{black}{\textbf{$\textlangle$content type set$\textrangle$}} denotes set $\Phi^{\text{cont}}$. 

\section{D. Experimental Details}

\subsection{D.1. Datasets}
In this section, we introduce the datasets used in our study, starting with the specific parameters of the IMDB, DBLP, and ACM datasets. 
The parameters of the aforementioned datasets are described in Table \ref{table:dataset}. IMDB, DBLP, and ACM are commonly used heterogeneous graph datasets in the field of graph representation learning. These datasets each contain multiple types of nodes and edges, representing complex entity relationship networks. The IMDB dataset primarily involves relationships between movies, actors, and directors, and is applied in scenarios such as movie recommendation and character relationship analysis. The DBLP dataset focuses on academic publications, including node types such as papers, authors, conferences, and keywords, and is used for academic network analysis and recommendation system research. The ACM dataset is similar to DBLP but is centered on the field of computer science, involving node types like papers, authors, and research topics, and is utilized for studying academic influence, topic evolution, and collaboration across fields.

We further constructed two datasets, IMDB-RIR and DBLP-RID, based on the IMDB and DBLP datasets. IMDB-RIR was created by utilizing the node information from IMDB and conducting searches using the Google search engine. We collected a total of 3,043 node features, each comprising the top 10 search results from Google. These features were then used to replace the original data in the IMDB dataset, resulting in a more challenging heterogeneous dataset that closely resembles raw information from the internet. On the other hand, the DBLP-RID dataset was created by randomly deleting words from the existing node text data in DBLP, generating a noisier and more difficult-to-process dataset.

\begin{table*}\scriptsize
        \caption{Summary of datasets.}
	\begin{center}
		\begin{tabular}{lcccccc}
			\hline\rule{0pt}{5pt}
			
			{Name}  &  \#Nodes  & \#Node Types & \#Edges & \#Edge Types & Target & \#Classes  \\ 	
			\hline\rule{-3pt}{10pt}
			\text{IMDB} & {11616} & {3} &{34212} & {6} & {movie} & {3}\\
			\text{DBLP} & {26128} & {4} &{239566} & {6} & {author} & {4}\\
			\text{ACM} & {10942} & {4} &{547872} & {8} & {paper} & {3}\\

			\hline
		\end{tabular}
	\end{center}
        \label{table:dataset}
\end{table*}

\begin{table*}[t]\scriptsize
 	\caption{Summary of the hyperparameters used in each dataset. The zero value of $l^{\text{fmt}}$ indicates that since the node representations in the IMDB, DBLP, and ACM datasets are consistent, we did not apply any format adjustments in the alignment block.}
	\begin{center}
		\begin{tabular}{lccccccc}
			\hline\rule{0pt}{10pt}
			{Name}  &  Size of $\delta$ (MLP)  & Size of W & Size of $g$ (GNN) & $l^{\text{fmt}}$ & $l^{\text{cont}}$ & $L$ & $\alpha$ \\ 	
			\hline\rule{-3pt}{10pt}
			\text{IMDB} & {[768,256,3]} & {[(3,768,256),(3,256,256)]} & {[768,256,32]} & 0 & 2 & 2 & 0.7 \\ 
			\text{DBLP} & {[768,256,4]} & {[(4,768,256),(4,324,256),(4,128,256)]} & {[768,324,128,32]}  & 0 & 3 & 3 & 0.7  \\
			\text{ACM} & {[768,256,4]} & {[(4,768,512),(4,256,512)]} & {[768,256,32]} & 0 & 3 & 3 & 0.75 \\
                \text{IMDB-RIR} & {[768,256,3],[768,256,2]} & {[(3,768,256),(3,256,256)],[(2,768,64)]} & {[768,256,32]} & 1 & 2 & 2 & 0.7 \\
			\text{DBLP-RID} & {[768,256,4],[768,256,2]} & {[(4,768,256),(4,324,256),(4,128,256)],[(2,768,64)]} & {[768,324,128,32]} & 1 & 3 & 3 & 0.7 \\
			\hline
		\end{tabular}
	\end{center}
        \label{table:backbone}
\end{table*}

\subsection{D.2. Hyperparameters and Environments}
We have summarized the hyperparameters used for each of the different datasets in Table \ref{table:backbone}. All our experiments were conducted on a workstation with eight Quadro RTX 5000 GPU (16 GB), one Intel Xeon E5-1650 CPU, 128GB RAM, and a Unbuntu 20.04 operating system. 

\section{E. Further Experiments}

\subsection{E.1. Feature Visualization}

\begin{figure}[ht]
    \centering
    \subfigure[Input.]{%
        \includegraphics[width=0.13\textwidth]{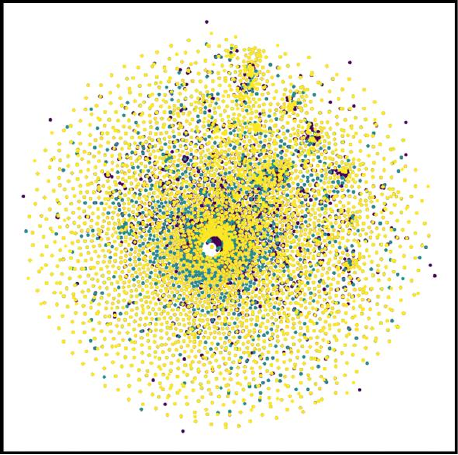}
        \label{fig:apsubfig1}
    }
    \subfigure[Output of LLM Processing.]{%
        \includegraphics[width=0.13\textwidth]{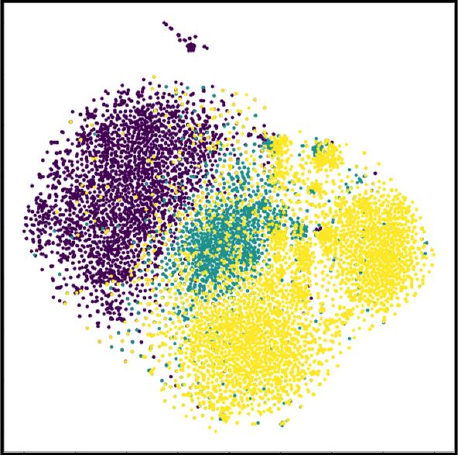}
        \label{fig:apsubfig2}
    }
    \subfigure[Output of the whole model.]{%
        \includegraphics[width=0.13\textwidth]{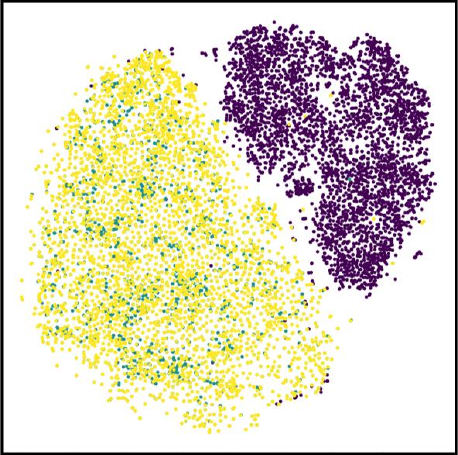}
        \label{fig:apsubfig3}
    }
    \\
    \subfigure[Input.]{%
        \includegraphics[width=0.13\textwidth]{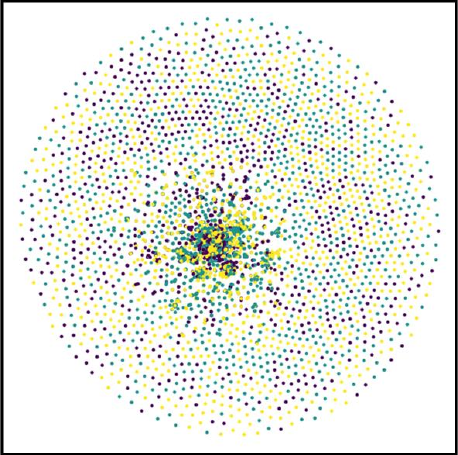}
        \label{fig:apsubfig4}
    }
    \subfigure[Output of LLM Processing.]{%
        \includegraphics[width=0.13\textwidth]{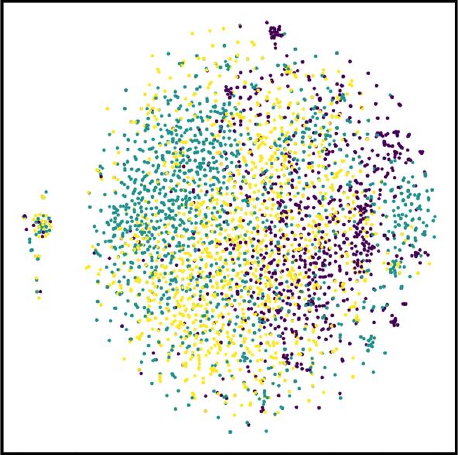}
        \label{fig:apsubfig5}
    }
    \subfigure[Output of the whole model.]{%
        \includegraphics[width=0.13\textwidth]{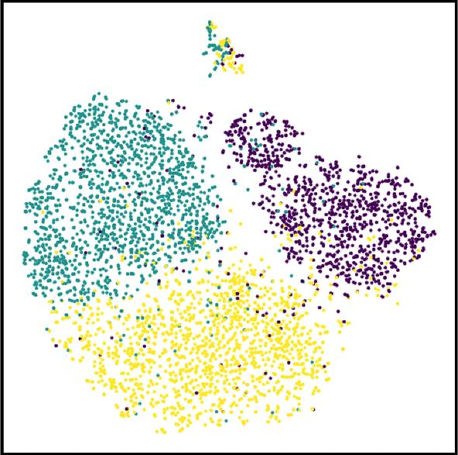}
        \label{fig:apsubfig6}
    }
    \caption{The IMDB data representations at each stage of the model, after dimensionality reduction using the t-SNE method. Different colors represent different type((a),(b),(c)) or class((d),(e),(f)) of nodes.}
    \label{fig:acm}
\end{figure}

\begin{table}[ht]\scriptsize
    \renewcommand{\arraystretch}{1.3}
    \setlength{\tabcolsep}{5pt}
    \centering
    \caption{Ground-truth type experiment results.}
    \begin{tabular}{l|cc|cc|cc}
        \hline
        Datasets & \multicolumn{2}{c}{IMDB}  & \multicolumn{2}{c}{DBLP} & \multicolumn{2}{c}{ACM}\\
        \cline{1-7}
        Metrics & Macro-F1 & Micro-F1  & Macro-F1 & Micro-F1 & Macro-F1 & Micro-F1\\
        \hline
        GHGRL & 70.34$\pm$0.60 &70.61$\pm$0.26  & 90.85$\pm$0.50 &91.44$\pm$0.76 & 92.85$\pm$0.22 &92.78$\pm$0.59\\
        \hline
        GHGRL with Ground-truth type & \textbf{70.42$\pm$0.34} & \textbf{70.82$\pm$0.80} & \textbf{90.91$\pm$0.30} & \textbf{91.53$\pm$0.71} & \textbf{92.94$\pm$0.22} & \textbf{92.93$\pm$0.56J}\\
        \hline
    \end{tabular}
    \label{tab:true_label_experiments}
    \vskip -0.1in
\end{table}

\begin{table}[ht]\scriptsize
    \renewcommand{\arraystretch}{1.3}
    \setlength{\tabcolsep}{5pt}
    \centering
    \caption{Ablation experiment results.}
    \begin{tabular}{l|cc|cc|cc}
        \hline
        Datasets & \multicolumn{2}{c}{IMDB (20\% Training)} & \multicolumn{2}{c}{IMDB (40\% Training)} & \multicolumn{2}{c}{IMDB (80\% Training)}\\
        \cline{1-7}
        Metrics & Macro-F1 & Micro-F1 & Macro-F1 & Micro-F1& Macro-F1 & Micro-F1\\
        \hline
        GCN   & 59.53$\pm$0.56 & 59.81$\pm$0.13 & 60.13$\pm$0.76 & 60.38$\pm$1.19 & 62.14$\pm$0.16 & 61.80$\pm$0.32\\
        \hline
        GHGRL without confidence score & 69.44$\pm$0.23  & 69.73$\pm$0.53 & 70.41$\pm$0.40 & 70.59$\pm$0.43 & 72.51$\pm$0.23 & 72.66$\pm$0.18\\
        GHGRL without $\bm{h}^{\text{reas}}_{v}$  & 69.68$\pm$0.85  & 70.03$\pm$0.63  &71.24$\pm$0.43 &71.52$\pm$0.40 &73.03$\pm$0.22 &73.36$\pm$0.36\\
        \hline
        GHGRL & \textbf{70.34$\pm$0.60} & \textbf{70.61$\pm$0.26} & \textbf{72.13$\pm$0.64} & \textbf{72.46$\pm$0.62} & \textbf{76.38$\pm$0.51} & \textbf{76.64$\pm$0.47}\\
        \hline
    \end{tabular}
    \label{tab:ablation_experiments}
    \vskip -0.1in
\end{table}

\begin{table}[h!]\scriptsize
\centering
\begin{tabular}{c|>{\columncolor{cyan!20}}p{4.5cm}|c|>{\columncolor{cyan!20}}p{4.5cm}|c}
\hline
\textbf{Node Feature} & \textbf{Description $h^{\text{desc}}$} & \textbf{Estimated Type Name} & \textbf{Reason $h^{\text{reas}}$} & \textbf{Confidence Score} \\ \hline
Avatar & "Avatar" is a 2009 science fiction film directed by James Cameron, set in the future on the planet Pandora, where a paraplegic marine named Jake Sully becomes an avatar of a native Na'vi warrior, leading to a clash between the human corporation and the native tribe over the mining of a valuable mineral. & movie & The result in question 1 can have a significant impact on the result in question 2, as it provides a set of keywords that can be used to identify the plot and themes of the movie, which can inform the selection of movie directors or actors. & 0.8 \\ \hline
Óscar Jaenada & Óscar Jaenada is a Spanish actor known for his roles in films such as "Life" (2017), "The Shallows" (2016), and "Instructions Not Included" (2013), as well as his work in Spanish television and theater. & movie actors & The result of this description can have an impact on the result of this being classified as a "movie actor" by providing more information about the actor's career and abilities. & 0.9 \\ \hline
Agnieszka Holland & A description of the given content is that it is the name of a person, specifically a film director, Agnieszka Holland. & movie directors & Knowing the name of a film director can have an impact on the result of identifying the movies they have directed, as it provides a specific individual to research and associate with a particular film or genre. & 0.9 \\ \hline
Takeo Kanade & Takeo Kanade is a renowned computer scientist and robotics engineer who has made significant contributions to the fields of computer vision, robotics, and artificial intelligence. He is known for his work on facial recognition, object recognition, and human-computer interaction, among other areas. & Authors & The result of identifying the author's name can have an impact on the credibility and authority of the research paper or publication, as it provides information about the expertise and qualifications of the researcher. & 0.8 \\ \hline
AAAI & The content refers to the Association for the Advancement of Artificial Intelligence, a prominent organization in the field of artificial intelligence that hosts conferences and publishes papers. & Feature: conferences & The result of describing the content can have an impact on the identification of the conference type, as knowing the name of the conference can help to categorize it as a specific type of conference, such as a machine learning or natural language processing conference. & 0.8 \\ \hline
automatic & The given content is a term used to describe a computer system that can perform tasks without human intervention, such as processing and analyzing data, making decisions, and taking actions, all without the need for human oversight or control. & Feature: Keywords & The result of describing the content as "automatic" can have a significant impact on the feature "keywords" as it can help identify relevant and important terms that are related to the concept of automation in computer science. & 1.0 \\ \hline
\end{tabular}
\caption{Demonstration of the correct answers of the LLM.}
\label{tab:content_tabler}
\end{table}

\begin{table}[h!]\scriptsize
\centering
\begin{tabular}{c|>{\columncolor{red!20}}p{4.5cm}|c|>{\columncolor{red!20}}p{4.5cm}|c}
\hline
\textbf{Node Feature} & \textbf{Description $h^{\text{desc}}$} & \textbf{Estimated Type Name} & \textbf{Reason $h^{\text{reas}}$} & \textbf{Confidence Score} \\ \hline
Wagner Truppel & Wagner Truppel is a type of paper format or layout used in academic publishing, particularly in computer science and related fields. It is a specific way of organizing and presenting the content of a research paper, including the title, authors, abstract, introduction, methodology, results, discussion, and conclusion. & papers title & The impact of using the Wagner Truppel format can be that it helps to improve the clarity and readability of the paper, making it easier for readers to follow the author's argument and understand the research findings. This, in turn, can increase the paper's chances of being accepted for publication and cited by other researchers. & 0.8 \\ \hline
\end{tabular}
\caption{Demonstration of the wrong answers of the LLM.}
\label{tab:content_tablew}
\end{table}


We also utilized the t-SNE method to visualize the node features of our model at different stages using the IMDB dataset, as illustrated in Figure \ref{fig:acm}. This process mirrors the observations made with the ACM dataset. Initially, at the input stage, nodes of different types and classes are disordered and lack a clear structure. However, after processing by the LLM, these features begin to differentiate, with different types of nodes becoming highly intertwined, forming several distinct clusters. Despite this, the different classes of paper nodes remain intermingled and challenging to distinguish. Ultimately, after the data is processed by PAGNN, our model successfully organizes the nodes into three clearly separated groups based on their types, with the paper nodes distinctly categorized into three classes. This outcome indicates that PAGNN further enhances the information produced by the LLM, resulting in a more refined and accurate classification.

\subsection{E.2. Ground-Truth Type Experiment}

We evaluated the performance of using ground-truth types versus estimated types generated by the LLM, as shown in Table \ref{tab:true_label_experiments}. The results indicate that the method using ground-truth types slightly outperforms the method using estimated types. However, given the proportion of correctly estimated nodes discussed earlier, we can conclude that the estimated types generated by the LLM are sufficiently accurate. Additionally, the confidence module helps mitigate the impact of classification errors in the types produced by the LLM. Overall, the quality of node features, rather than the accuracy of estimated types, is the key factor influencing the results.

\subsection{E.3. Ablation Study}

We conducted comprehensive ablation studies by removing two modules separately: (1) Removing the confidence module, where parameters were selected solely based on estimated types generated by the LLM, without using confidence levels to assign different proportions of parameters; (2) Removing $\bm{h}^{\text{reas}}_{v}$ from the LLM-generated answers. $\bm{h}^{\text{reas}}_{v}$ is concatenated with the estimated types generated by the LLM to enhance this content.
Results are shown in Table \ref{tab:ablation_experiments}. These results demonstrate the utilization of these two modules.

We conducted comprehensive ablation studies by removing two modules separately: (1) the confidence module, where parameters were selected solely based on the estimated types generated by the LLM, without using confidence levels to assign different proportions of parameters; and (2) the removal of $\bm{h}^{\text{reas}}{v}$ from the LLM-generated outputs. $\bm{h}^{\text{reas}}{v}$ is typically concatenated with the estimated types generated by the LLM to enhance the content. The results, presented in Table \ref{tab:ablation_experiments}, demonstrate the importance of these two modules in the overall model performance.

\subsection{E.3. Illustration of the LLM Answers}
In Tables \ref{tab:content_tabler} and \ref{tab:content_tablew}, we present examples of correct and incorrect responses from the LLM, respectively, to facilitate a more thorough analysis of the model. It can be observed that, regardless of the type of response, the LLM is capable of providing additional background knowledge for the nodes based on its inherent capabilities. Our approach specifically encourages this by modifying the prompt.

\end{document}